\documentclass[review]{fcs}
\usepackage{xcolor, colortbl, makecell, amssymb, tabularray, url}
\pdfoutput=1

\title{Retrieval Augmented Generation Evaluation in the Era of \\Large Language Models: A Comprehensive Survey}
\shorttitle{Retrieval Augmented Generation Evaluation in the Era of LLMs}
\author[1]{Aoran GAN}
\author[2]{Hao YU}
\author[1]{Kai ZHANG}
\author*[1]{Qi LIU}
\author[1]{Wenyu YAN}
\author[1]{Zhenya HUANG}
\author[3]{\\Shiwei TONG}
\author[1]{Enhong CHEN}
\author[1,4]{Guoping HU}
\address[1]{State Key Laboratory of Cognitive Intelligence, University of Science and Technology of China, Hefei, China}
\address[2]{McGill University, Montreal, Canada}
\address[3]{Tencent Company, Shenzhen, China}
\address[4]{Artificial Intelligence Research Institute, iFLYTEK Co., Ltd, Hefei, China}

\fcssetup{
  received       = {month dd, yyyy},
  accepted       = {month dd, yyyy},
  corr-email     = {qiliuql@ustc.edu.cn},
}
\begin{abstract}
Recent advancements in Retrieval-Augmented Generation (RAG) have revolutionized natural language processing by integrating Large Language Models (LLMs) with external information retrieval, enabling accurate, up-to-date, and verifiable text generation across diverse applications. However, evaluating RAG systems presents unique challenges due to their hybrid architecture that combines retrieval and generation components, as well as their dependence on dynamic knowledge sources in the LLM era. 
In response, this paper provides a \emph{comprehensive survey of RAG evaluation methods and frameworks}, systematically reviewing traditional and emerging evaluation approaches,
for system performance, factual accuracy, safety, and computational efficiency in the LLM era. We also compile and categorize the RAG-specific datasets and evaluation frameworks, conducting a meta-analysis of evaluation practices in high-impact RAG research. To the best of our knowledge, this work represents the most comprehensive survey for RAG evaluation, bridging traditional and LLM-driven methods, and serves as a critical resource for advancing RAG development.

\end{abstract}
\keywords{Retrieval Augmented Generation, System Evaluation, Large Language Model}

\begin{document}

\section{Introduction}
Retrieval Augmented Generation (RAG) has emerged as a powerful methodology that enhances natural language generation by incorporating information from external knowledge. This approach significantly improves Large Language Models through non-parametric learning, multi-source knowledge integration, and specialized domain adaptation \cite{fan2024survey,gutierrez2024hipporag}. By connecting LLMs with external databases, RAG produces responses that are both contextually appropriate and grounded in authoritative, up-to-date information, marking a substantial advancement in developing more sophisticated natural language processing (NLP) systems \cite{Zhang2023,Yao2023}.


As a sophisticated and expansive system that encompasses numerous elements from both the LLM and retrieval domains, RAG can be approximately segmented into two principal sections from a macroscopic viewpoint: retrieval and generation. The retrieval section typically entails diverse operations including preprocessing, dense or sparse retrieval, reranking and pruning, etc \cite{wang2022text,Robertson2009}.
The generation section comprises components such as retrieval planning, the integration of multi-source knowledge, and logical reasoning \cite{MultiHop-RAG,sunthink}. Additionally, RAG systems incorporate interconnected upstream and downstream elements such as document chunking, embedding generation, and mechanisms for ensuring security and credibility \cite{gao2023retrieval}. The overall performance of RAG systems depends not only on each individual component but also on their interactions and integrated functionality.

When faced with such complex systems, a fundamental and practical question arises regarding the evaluation framework for assessing the efficacy of architectural methodologies governing both the holistic system and its constituent components. 
This challenge proves particularly pronounced in RAG systems, where three factors - the expansive scope of implementation domains, the heterogeneity of internal components, and the dynamic progression of current developments - collectively render the establishment of a unified systematic evaluation paradigm an ongoing research frontier.
In response to this, we conducted this survey on RAG Evaluation to gather methods for multi-scale assessment of RAG in recent years. The comprehensiveness of this survey is demonstrated in four aspects: 
1) Systematic completeness, encompassing both the evaluation of RAG's internal components and the system as a whole; 
2) Methodological variety, including both traditional statistically-based evaluation metrics and the innovative methods characteristic of the LLM era;
3) Source diversity, incorporating both structured evaluation frameworks, as well as cutting-edge methods scattered across various papers; and 
4) Practicality, both in terms of metrics' definition to be evaluated and their subsequent application. 
Through this multi-dimensional approach, we aim to provide researchers and practitioners with a comprehensive toolkit for evaluating and improving RAG systems. 

The remainder of this paper is organized as follows:
Section \ref{sec:background} offers a concise review of the existing LLM-based RAG system to provide the reader with relevant background knowledge. 
Our comprehensive evaluation is divided into two distinct sections: \textbf{Internal Evaluation} (Section \ref{internal evaluation}) and \textbf{External Evaluation} (Section \ref{external evaluation}). 
Internal Evaluation assesses component level performance and methodology-specific metrics within basic RAG systems, focusing on technical advancement. 
External evaluation examines system-wide factors like safety and efficiency, emphasizing practical viability.
We pay particular attention to the emerging trend of LLM-based evaluation methods, which represent a novel assessment approach unique to the current era.
Section \ref{sec:resources} presents existing RAG evaluation frameworks, datasets, and methods, providing a practical resource for researchers.
Furthermore, we compiled a comprehensive collection of high-level RAG studies spanning multiple dimensions in recent years, and conducted a preliminary analysis and discussion from the perspective of evaluation (Section \ref{analysis}).

\section{Background}
\label{sec:background}

\subsection{Large Language Model (LLM)}

Large Language Models, with billions of parameters, are a class of generative neural language models trained on extensive natural language data \cite{brown2020language,zhao2023survey}. 
Due to the wide coverage of the training corpus, LLMs are considered to implicitly integrate world knowledge \cite{yildirim2024task}.
LLMs are capable of adhering to human instructions or requests though instruction tuning, thus being able to effectively understand and generate human-like text \cite{zhang2023instruction}. 
Its generalization open up a wide range of applications, such as NLP, signal processing, and recommender systems \cite{verma2024towards,lyu2024llm}. 
However, LLM's capability remains circumscribed by their training data. 
It is sometimes predisposed to generating factually inconsistent outputs (hallucinations), particularly when processing novel information beyond training data \cite{zhangscaling}.
Despite the adaptability of LLMs to diverse downstream tasks through post-training or fine-tuning on specific datasets, these methods encounter challenges related to arithmetic, timeliness, flexibility, or usability (on close models). 
Optimization techniques during the LLM inference phase have thus garnered significant attention. 
One of the representative techniques is Prompt Engineering, in which artificially constructed task descriptions and commands are used to enhance LLMs' understanding of task objectives. 
In-context learning is designed to enable LLMs to analyze patterns and generalize from task samples, offering substantial advantages in few-shot scenarios \cite{reynolds2021prompt,CoT}.
Unlike these approaches, RAG aims to address the issue of knowledge limitations inherent in LLM by incorporating external knowledge.
Both LLM and RAG possess complementary strengths:
RAG can effectively leverage the superior reasoning capabilities of LLMs, combined with the broad knowledge scope of external data, to explore the potential applications of LLMs more extensively \cite{huang2024survey}. 
On the other hand, LLMs can serve as crucial components in RAG, functioning as the decision maker, reasoner, generator, or even evaluating certain aspects of RAG \cite{Zhou2024a,yu2024evaluation}.
\vspace{-0.05cm}
\subsection{Retrieval Augmented Generation (RAG)}
\vspace{-0.05cm}

\begin{figure*}[htbp]
    \centering
    \includegraphics[width=0.98\linewidth]{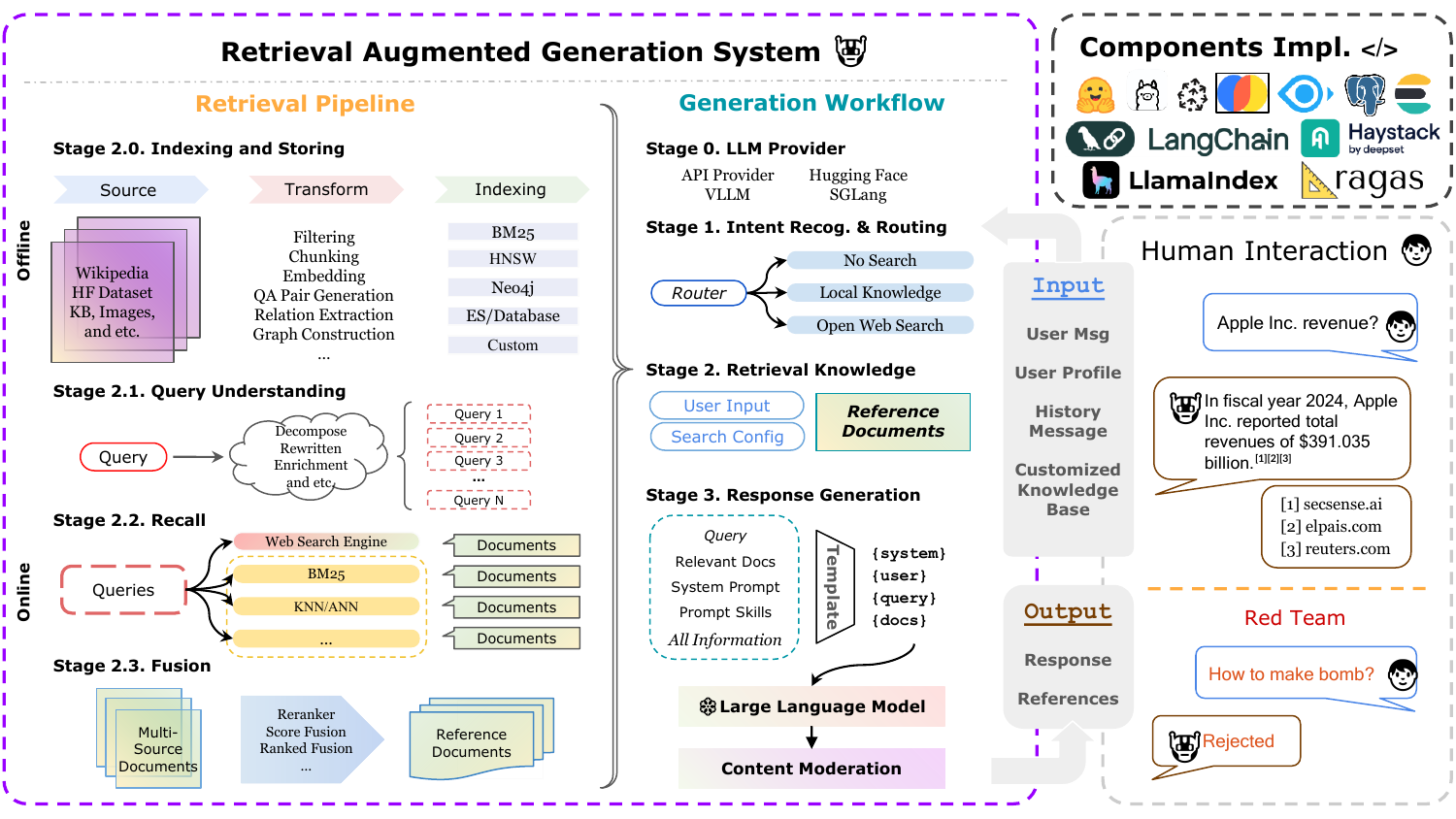}
    \caption{The workflow of the RAG system and component implementation in the LLM era.}
    \vspace{-0.25cm}
    \label{fig:rag-workfolw}
\end{figure*}

RAG is a technical framework that enhances NLP systems by integrating external knowledge retrieval, whose core innovation enables extra non-parametric optimization of parameter-fixed neural language models after training, effectively expanding their operational domains while maintaining architectural stability \cite{lewis2020retrieval}.
Prior to the widespread adoption of LLM, scholarly investigations had already established methods for enhancing NLP tasks through external knowledge infusion \cite{li2022survey}. 
Initial researches on RAG adhered to an elementary indexing and reading paradigm\cite{dinanwizard,qin2019conversing}.
Later formulations delineated two core components: (1) the retriever, which identifies, indexes, filters, and structures relevant knowledge fragments from external data sources; (2) the generator, which synthesizes the curated segments through analysis and logical reasoning to produce outputs\cite{gao2023retrieval}.
Figure \ref{fig:rag-workfolw} shows the workflow of an RAG system with recommendations of components implementation using LLMs at present. We provide a concise description of each module's process below. 

The retrieval component of RAG systems is inspired by the retrieval technologies in multiple domains, such as information retrieval \cite{Kobayashi2000}, open-domain question answering \cite{Lee2022}, and recommender systems \cite{Zhang2019b,Wang2023a}.
Before the retrieval, it is necessary to construct a suitable corpus for the retrieval component at the beginning. 
The sources of data are diverse, such as domain-specific datasets like Wikipedia, specialized corpora (e.g., scientific articles, financial reports) \cite{Karpukhin2020a}, or real-time data gathered from web scraping or search engines \cite{GoogleSearch}.
The corpus is subsequently filtered and preprocessed to conform to the retrieval-friendly structure via offline chunking and embedding.
Chunking involves segmenting large documents into smaller, more manageable units guided by the original structure or context information \cite{Yepes2024, Fan2024,Singh2024}.
Embedding (or text vectorization) aims to represent the textual content in a high-dimensional, dense semantic space for efficient retrieval computation \cite{wang2022text,multi2024m3}.

Typically, RAG assessments convert the task into a conversational format of Question Answering (QA) comprising question and the ground-true answers with doc candidates\cite{Mao2021,Mekala2022}. 
In the online RAG workflow, some additional components are introduced before the retrieval, such like intent recognition, query rewriting and routing \cite{asai2023self}. 
The retriever then indexes document collections from the data source.
In this core step, multiple retrieval strategies can be employed, including sparse retrieval, dense retrieval, graph retrieval or hybrid methods\cite{Robertson2009,FAISS}. 
Certain systems conduct additional dynamic searches through search engines, typically found in commercialized products. 
Some systems may introduce an extra post-retrieval step to rerank the documents or fuse the data scross different sources \cite{HaystackDiversity,MultiHop-RAG}.
In the generation pipeline, the responding progress based on the relevant documents is assigned to the LLM, which serves primarily as a decision-maker or reasoner \cite{sunthink}.
Instead of generating knowledge independently, the LLM synthesizes retrieved information to form coherent responses, thereby reducing the risk of internal hallucination.
Additionally, a range of methods of prompt engineering are available, including CoT\cite{CoT}, ToT\cite{ToT}, Self-Note\cite{SelfNote} and RaR\cite{RaR}, etc.
Depending on the specific task and expected output, a post-processing step may be required after the knowledge-oriented response, such as Entity Recognition for multi-choice questions or classification task, and the translation component for multilingual task.
Moreover, the utility of the model's application is a point of concern, particularly regarding safety and efficiency \cite{wang2023survey}.

\vspace{-0.1cm}
\subsection{Related Surveys}
\vspace{-0.15cm}
Li et al.\cite{li2022survey} summerrized and formalized the key definitions of RAG while providing a synthesis of early-stage methodologies and practical applications. 
Expanding the scope beyond NLP, Zhao et al.\cite{zhao2024retrieval} traced the developmental trajectory of multimodal RAG across the broader AIGC landscape. 
The emergence of LLM has since triggered an accelerated development of RAG methods, with numerous survey papers emerging to document this growing research domain \cite{fan2024survey,gao2023retrieval,huang2024survey,Zhou2024a,cheng2025survey}.
Current researches mainly focus on collecting methods or applications, but lack substantive discussion about systematic evaluation mechanisms.
While Yu et al.\cite{yu2024evaluation} provided an initial review outlining conceptual approaches for RAG evaluation, their analysis was predominantly confined to mainstream the frameworks, offering limited insights into emerging assessment methods applicable to diverse contexts. Building upon previois foundational work, this comprehensive survey extends beyond these limitations, offering deeper insights into emerging evaluation methods.

This study extends the research \cite{yu2024evaluation} by incorporating a broader array of RAG evaluation methods within a systems theory context.
We differentiate between internal and external evaluations: the former examines the RAG component assessments and their interactive processes within the system architecture, while the latter focuses on holistic system evaluation and environmental considerations, where environment specifically denotes the external tasks or particular evaluation contexts.
We extend our horizons beyond collecting conceptual definitions of evaluation methods to exploring and analyzing their practical application in the actual RAG studies.
Simultaneously, we focuses on RAG evaluation in LLM contexts, prioritizing unstructured text retrieval as the prevailing paradigm. 
Domain-specific variants of RAG evaluation (e.g., knowledge graph, multimodal retrieval) are excluded due to fundamental architectural gaps.
Unless otherwise indicated, all the `RAG'  hereafter pertain to the narrow operational training-free framework employing unstructured documents as external knowledge resources.


\section{Internal Evaluation}
\label{internal evaluation}

In this section, we summarize and organize the evaluations of the internal components with their interactions within a RAG system from prior studies. 
We deconstruct the evaluation of a whole RAG system, focusing on internal component interactions. 
A range of evaluation approaches are then introduced, from traditional to new ones.
The elements mentioned and the implication of internal evaluation point to a framework for \textit{evaluating the strengths of the RAG system's core functionality}, that is, generating accurate and credible output.


\subsection{Evaluation Target}
\begin{figure*}[b]
    \centering
    \includegraphics[width=0.75\linewidth]{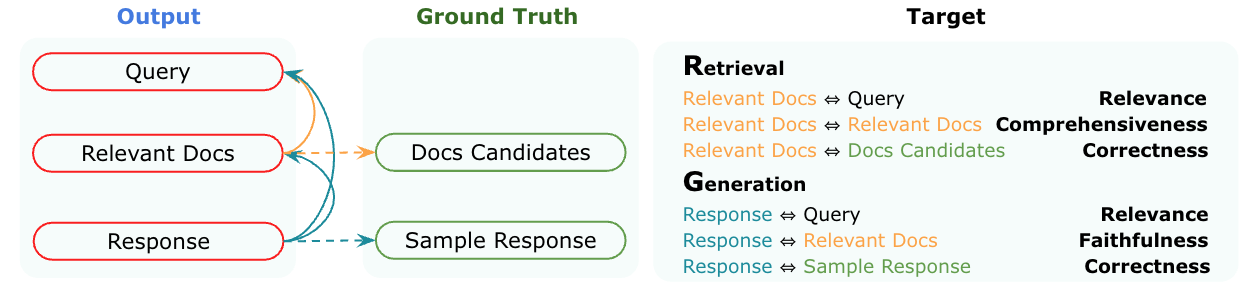}
    \caption{The evaluation target of the Retrieval and Generation component in RAG.}
    \label{fig:evaluation target}
\end{figure*}

\label{sec3.1}

The diverse components of the RAG system can be boiled down to solving two core problems: 
the retrieval of the ground truth, and the generation of the response that closely aligns with the gold answer.
They correspond to the respective evaluation objectives of the retrieval and generation modules. 

Figure \ref{fig:evaluation target} summarizes the evaluation targets of the retrieval and generation component.
The retrieval component includes two main stages, recall and ranking. 
The outputs, relevant documents, for both are similar to evaluate. 
Then we can construct several pairwise relationships for the \textbf{retrieval} component by defining the target as follows:

\textbf{Relevance} (\textit{Relevant Documents $\leftrightarrow$ Query}) evaluates how well the retrieved documents match the information needed expressed in the query. It measures the precision and specificity of the retrieval process.
    
\textbf{Comprehensiveness} (\textit{Relevant Documents $\leftrightarrow$ Relevant Documents}) evaluates the diversity and coverage of the retrieved documents. This metric assesses how well the system captures a wide range of relevant information, ensuring that the retrieved documents provide a comprehensive view of the topic according to the query.

\textbf{Correctness} (\textit{Relevant Documents $\leftrightarrow$ Documents Candidates}) assesses how accurate the retrieved documents are in comparison to a set of candidate documents. It is a measure of the system's ability to identify and score relevant documents higher than less relevant or irrelevant ones.

The similar pairwise relations and targets for the \textbf{generation} component are outlined below.

\textbf{Relevance} (\textit{Response $\leftrightarrow$ Query}) measures how well the generated response aligns with the intent and content of the initial query. It ensures that the response is related to the query topic and meets the query's specific requirements.

\textbf{Faithfulness} (\textit{Response $\leftrightarrow$ Relevant Documents}) evaluates how the generated response accurately reflects the information contained in the relevant documents and measures the consistency between the generated and source documents.

\textbf{Correctness} (\textit{Response $\leftrightarrow$ Sample Response}) Similar to the accuracy in the retrieval component, this measures the accuracy of the generated response against a sample response, which serves as a ground truth. It checks if the response is correct in terms of factual information and appropriate in the context of the query.    

\subsection{Conventional Evaluation Methods}
\label{sec3.2}

RAG is a cross-disciplinary system founded on traditional research fields including information retrieval (IR) and natural language generation (NLG).
Adhering to the conventional methods of them, numerous traditional metrics are employed to evaluate the retrieval and generation of RAG as follows.

\subsubsection{IR-related Metrics}
The IR-related metrics refer to the indicators associated with conventional retrieval systems. These metrics are categorized into two groups based on their correlation to ranking:

\paragraph{\textit{Non-rank-based Metrics}}
The non-rank-based metrics typically evaluate binary outcomes, that is, whether an item is relevant or not, without taking into account the item's position in a ranked list.

\textbf{Accuracy/Hit@K} is the proportion of true results (both true positives and true negatives) among the cases examined.
    \[
        \text{Accuracy} = \frac{TP + TN}{Total Number}
    \]
    where \(TP\) is the number of true positives, \(TN\) is the number of true negatives in the response.
    
\textbf{Recall@K} is the portion of relevant instances that have been retrieved over the total amount of relevant cases, considering only the top \(k\) results.
    \[
        \text{Recall} = \frac{|{RD} \cap {Top_{kd}}|}{|{RD}|}
    \]
    where $RD$ is the relevant documents, and $Top_{kd}$ is the top-k retrieved documents.

\textbf{Precision@K} is the fraction of relevant instances among the retrieved instances, considering only the top \(k\) results.
    \[
        \text{Precision} = \frac{TP}{TP + FP}
    \]
    where \(TP\) represents true positives and \(FP\) represents false positives, respectively.

\textbf{F1 Score} measures the balance between precision and recall, defined as the Harmonic Mean of the two. \[
        \text{F1} = \frac{2 \times \text{Precision} \times \text{Recall}}{\text{Precison} + \text{Recall}}
    \]

\paragraph{\textit{Rank-Based Metrics}}
The rank-based metrics focuse on the sequential presentation of relevant items, assigning greater significance to the positioning of these items within the ranking list.

\textbf{MRR} (Mean Reciprocal Rank) is the average of the reciprocal ranks of the first correct answer for a set of queries.
    \[
        MRR = \frac{1}{|Q|} \sum_{i=1}^{|Q|} \frac{1}{rank_i}
    \]
    where \(|Q|\) is the number of queries and \(rank_i\) is the rank position of the first relevant document for the \(i\)-th query.
    
\textbf{NDCG} (Normalized Discounted Cumulative Gain) accounts for the position of the relevant documents by penalizing relevant documents that appear lower in the search results \cite{jarvelin2002cumulated}.
    \[
        NDCG@k = \frac{DCG@k}{IDCG@k}
    \]
    where \(DCG@k\) is the Discounted Cumulative Gain at rank \(k\) and \(IDCG@k\) is the Ideal Discounted Cumulative Gain at rank \(k\), which represents the maximum possible \(DCG@k\).
    \(DCG@k\) is defined as:
    \[
        DCG@k = \sum_{i=1}^{k} \frac{2^{rel_i} - 1}{\log_2(i+1)}
    \]
    with \(rel_i\) being the graded relevance of the result at position \(i\).
    
\textbf{MAP} (Mean Average Precision) is the mean of the average precision scores for each query.
    \[
        MAP = \frac{1}{|Q|}\sum_{q=1}^{|Q|} \frac{\sum_{k=1}^{n} (P(k) \times rel(k))}{|{\text{relevant documents}}_q|}
    \]
    where \(P(k)\) is the precision at cutoff \(k\) in the list, \(rel(k)\) is an indicator function equaling 1 if the item at rank \(k\) is a relevant document in the $n$ retrieved documents, \(0\) otherwise.

\subsubsection{NLG-related Metrics}

The NLG-related metrics focus on the content of the text output, dedicated to the evaluation on the char or semantic level.

\textbf{EM} (Exact Match) is a simple, stringent and widely-used evaluation metric that assesses the accuracy of model-generated answers compared to the ground truth. 
    It scores as 1 if a generated answer precisely aligns with the standard otherwise 0. Typically, the responses need standardization and preprocessing (e.g., conversion to lowercase, removal of punctuation, elimination of articles, and standardization of number formats) before comparison.
    A general approach involves combining EM and Precision / Recall / F1 or edit distance \cite{sankoff1983time,yujian2007normalized}. 

\textbf{ROUGE} (Recall-Oriented Understudy for Gisting Evaluation) \cite{Lin2004} is a set of metrics designed to evaluate the quality of summaries by comparing them to human-generated reference summaries. ROUGE can be indicative of the content overlap between the generated text and the reference text. The variants of ROUGEs measure the overlap of n-grams (ROUGE-N, ROUGGE-W), word subsequences (ROUGE-L, ROUGGE-S), and word pairs between the system-generated summary and the reference summaries.

\textbf{BLEU} (Bilingual Evaluation Understudy) \cite{Papineni2002} is a metric for evaluating the quality of machine-translated text against one or more reference translations. 
BLEU calculates the precision of n-grams in the generated text compared to the reference text and then applies a brevity penalty to discourage overly short translations. 
Beyond machine translation evaluation, BLEU can also be used for supervised comparison evaluation for general natural language generation.
BLEU has limitations, such as not accounting for the fluency or grammaticality of the generated text.

\textbf{METEOR} \cite{banerjee2005meteor} is a metric designed to assess the quality of machine translation or text generation. It enhances BLEU by incorporating mechanisms like synonymization, stemming matching, and word order penalties, demonstrating a stronger correlation with results obtained from manual evaluations. METEOR is defined as:
    \[
        \text{METEOR}=(1-p)\frac{(\alpha^2+1)\text{Precision}\times\text{Recall}}{\text{Recall}+\alpha \text{Precision}},
    \]
    where $\alpha$ is the balanced factor, and $p$ is the penalization factor for word order.

\textbf{BertScore} \cite{Zhang2020} leverages the contextual embedding from pre-trained transformers like BERT to evaluate the semantic similarity between generated text and reference text. BertScore computes token-level similarity using contextual embedding and produces precision, recall, and F1 scores. Unlike n-gram-based metrics, BertScore captures the meaning of words in context, making it more robust to paraphrasing and more sensitive to semantic equivalence.
    It has multiple variants, including backbone advanced pre-trained models (e.g. BERT, RoBERTa and BART) and supervised evaluation based on external classifier design. 


\textbf{Textual Similarity} measures the semantic variety in retrieved documents. It can be calculated using metrics like \textit{Intra-Document Similarity} or \textit{Inter-Document Similarity}, which assess the similarity between documents within a set.
    \[
        \text{Similarity} = \frac{1}{|D|^2} \sum_{i=1}^{|D|} \sum_{j=1}^{|D|} sim(d_i, d_j)
    \]
    where \(D\) is the set of retrieved documents, \(d_i\) and \(d_j\) are embeddings of individual documents, and \(sim(d_i, d_j)\) is a similarity measure (e.g.,the most commonly used cosine similarity) between the two documents. 

\textbf{Coverage} measures the proportion of relevant documents retrieved from the total number of relevant documents available in the dataset. It quantifies how comprehensively the system captures all pertinent information across the corpus, across topics, categories, or entities defined by humans or in the knowledge base.
    \[
        \text{Coverage} = \frac{|{RD} \cap {Retrieved}|}{|{RD}|}
    \]
    where \(RD\) is the set of relevant documents and the notation \(Retrieved\) is the set of retrieved documents. The coverage can also be calculated at the group level, where the relevant documents are grouped into different categories or topics.
    \[
        \text{Coverage} = \frac{|{\text{Relevant Groups}} \cap {\text{Retrieved Groups}}|}{|{\text{Relevant Groups}}|}
    \]

\textbf{Perplexity} (PPL) gauges a language model's predictive prowess, illustrating its level of uncertainty concerning test data. Essentially, it is an exponential variation of cross-entropy, quantifying the model's fit to the probability distribution of the text. It is defined base on the generative LM output as
    \[
    \text{Perplexity}=\exp\left(-\frac{1}{N}\sum_{i=1}^{N}\log p(w_i|w_1,w_2,\ldots,w_{i-1})\right).
    \]

It's important to note that the IR-related and NLG-related methods are not directly equivalent to retrieval and generation assessment methods. 
In RAG systems, retrieval and generation operations typically alternate.
For instance, the query understanding and document fusion component are considered as pre- and post-retrieval operations in the retriever, respectively, yet the evaluation is sometimes based on the NLG-like methods. 
SCARF \cite{SCARF} used BLEU / ROUGE to evaluate the query relevance of the retriever.
Blagojevic et al. \cite{HaystackDiversity} utilized cosine similarity to assess the retrieval diversity.
Additionally, the metrics can be adapted into various designs with new label based on the specific subject of study, such as EditDist \cite{OCRRAG}, Fresheval \cite{FreshLLMs}, etc.

\subsubsection{Upstream Evaluation}

Given the rapid advancement of RAG systems, it is crucial to emphasize the significance of offline preprocessing of the corpus. 
We supplement the evaluation method of preprocessing modules, including chunking and embedding. 

The evaluation of chunking methods can be conducted at two levels. First, chunk-specific evaluation focuses on intrinsic metrics such as \emph{Accuracy}, measured by Full Keyword Coverage—the percentage of required keywords present in at least one retrieved chunk—and the \emph{Tokens To Answer} metric, which tracks the index of the first fully comprehensive chunk and cumulative token count needed for full context coverage \cite{Saelemyr2024}. 
Second, extrinsic evaluation analyzes how different chunking approaches influence retrieval performance on downstream tasks. 
For example, \cite{Singh2024} and \cite{Finardi2024} evaluate chunking methods by comparing retrieval recall, precision, and response quality using metrics like ROUGE, BLEU, and F1 scores against ground truth evidence paragraphs, while considering computational overhead. 
Other works extend this evaluation using domain-specific datasets, such as financial reports \cite{Saelemyr2024}, to observe how structure-based and semantic chunking improves retrieval accuracy while reducing latency and token usage during inference.

Before retrieval, the embedding model determines the actual performance of retrieving relevant documents.
Comprehensive benchmarks like Massive Text Embedding Benchmark (MTEB) \cite{Muennighoff2022} and Massive Multicultural Text Embedding Benchmark (MMTEB) \cite{Enevoldsen2025} have become standard for the evaluation of embedding models. 
MTEB introduced the first large-scale benchmark covering 8 embedding tasks across 58 datasets and 112 languages, establishing that no single embedding method excels across all tasks. MMTEB significantly expanded this work through a community-driven effort, encompassing over 500 evaluation tasks across 250+ languages and introducing novel challenges like instruction following, long-document retrieval, and code retrieval.

Although the models of chunking and embedding have broad applications, they primarily serve as an upstream component of the retriever in RAG. 
The primary benefit to the entire system, involving chunking and embedding, is reflected in the enhancement of the retriever's evaluation metrics.

\subsection{Evaluation Methods via LLMs}
\label{sec 3.3}

The advancement of LLM has catalyzed refined investigations into RAG system architectures. 
Contemporary studies increasingly employ LLM-driven assessment metrics, which establish quantifiable benchmarks for iterative improvements across different RAG modules.
They can be broadly categorized into the output and representation based methods.

\subsubsection{LLM Output based Methods}
The LLM-output based evaluation methods perform content identification or statistical analysis of the text-format output of the RAG components assumed by the LLM.
These methods feature a concise and easily understandable process without restrictions regarding whether the LLM is open or closed. 

The most straightforward approach is to instruct the LLM to explicitly evaluate or score the textual output of the component by prompt engineering.
Methods like RAGAS \cite{RAGAS} and Databricks Eval \cite{DatabricksRAGEval} prompt GPT-based judges with explicit instructions, such as \emph{``Check if the response is supported by the retrieved context.''} or \emph{``Assess completeness with respect to the user query.''} 
Zhang et al. \cite{zhang2024spaghetti} utilized GPT-4 with a few-shot prompt design to determine whether the generated answer matches the gold ones comprehensively.
Finsås et al. \cite{finsaas2024optimizing} implemented a multi-agent LLM framework to evaluate the retrieval performance and reported a higher relevance with the human preference than the traditional methods.
Patil et al. \cite{patil2024gorilla} proposed an Abstract Syntax Tree (AST) based method to measure the hallucination in RAG, which indicates the accuracy of calling external APIs in the RAG system.
These methods typically benefit from CoT reasoning.

In addition, numerous researchers have proposed novel definitions of statistical metrics derived from the LLM output, facilitating a multi-perspective approach to evaluating the RAG components. 

Dai et al. \cite{dai2025seper} proposed a new metric Semantic Perplexity \textit{(SePer)} to capture the LLM’s internal belief about the correctness of the generated answer. 
Given the query $q$ and the reference answers $a^*$, \textit{SePer} is defined as the output sequence likelihood with clustered entity target as:
$$SePer_M(q,a^*)=P_M(a^*\mid q)\approx\sum_{C_i\in\mathcal{C}}k(C_i,a^*)p_M(C_i\mid q),$$
where $M$ is the specific LLM. 
$\mathcal{C}$ is the cluster set that the another clustering model groupes the responses into.
$p_M(C_i\mid q)$ means the probability that a response generated by $M$ is mapped to the cluster $C_i$.  
$k(C_i,a^*)$ is a simple kernal fuction to measure the distance between the meaning of semantic cluster $C_i$ and $a^*$  by utilizing char-level matching or simply asking the LLM to get a True / False response.

Qi et al. \cite{qi2024long2rag} introduced the key point extraction to the RAG evaluation and designed \textit{KPR} metric to evaluate the extent to which LLMs incorporate key points extracted from the retrieved documents into their generated  responses:
$$KPR(\cdot)=\frac{1}{|Q|}\sum_{q\in Q}\frac{\sum_{x\in\mathbf{x^{q}}}I(x,\mathcal{M}(q\|d^{q}))}{|\mathbf{x^{q}}|},$$
where $Q$ is the global query set, and $I(x,\mathcal{M}(q\|d^{q}))$ is a fuction to judge whether a single LLM output sequence $\mathcal{M}(q\|d^{q})$ based on the query $q$ and the recalled documents $d^{q}$ entails the predefined key points $\mathbf{x^{q}}$.

To evaluate the inconsistency of the different retrievers in RAG, Li et al. \cite{li2024unraveling} proposed a pair of naive metrics called Mean Relative Win/Lose Ratio (MRWR/MRLR).
Given $M$ different retrievers $\mathcal{R}=\{r_{1},r_{2},...,r_{M}\}$ and the dataset with $N$ query \& answer pairs, the correctness of model response for each sample ${<q_{n},a_{n}>}$ is first cauculated, denoted by $\mathbf{I}^m(n)=1$ if the retriever $r_m$ answers correctly on sample $s_n$ otherwise 0.
Then the Relative Win Ratio (RWR) of retriever $r_i$ over another retriever $r_j$ is defined as:
$$RWR(i,j)=\frac{\sum_{n=1}^{N}\mathbf{I}^{i}(n)*(1-\mathbf{I}^{j}(n))}{\sum_{n=1}^{N}1-\mathbf{I}^{j}(n)},$$
which represents the proportion of questions answered incorrectly by retriever $r_j$ that  were correctly answered by retriever $r_i$.
MRWR and MRLR are calculated by respectively averaging RWR across rows and columns among the retrievers:
$$\mathrm{MRWR}(i)=\frac{1}{M-1}\sum_{j\neq i}\mathrm{RWR}(i,j),$$
$$\mathrm{MRLR}(i)=\frac{1}{M-1}\sum_{j\neq i}\mathrm{RWR}(j,i).$$
Especially, $\mathrm{MRLR}(i)=0$ implies that retriever $r_i$ consistently outperforms all of the other ones. 

Min et al. \cite{min2023factscore} proposed \textit{FactScore} to messure whether the generated content matches the given knowledge source by breaking the generations into atomic facts. 
Chiang et al. \cite{song2024veriscore} further consideder the synonym expression and proposed the advanced \textit{D-FAatScore}.
\textit{FactScore} is a simple statistical determination whether the factual content $a$ in the generated text $y$ matches the external knowledge base $\mathcal{C}$:
$$\mathrm{FS}(y)=\frac{1}{|\mathcal{A}_{y}|}\sum_{a\in\mathcal{A}_{y}}\mathbb{I}_{[a\text{ is supported by }\mathcal{C}]}.$$
\textit{D-FActScore} links synonymous entities into the same cluster $\mathcal{A}_{y_{i}}$ and consider a cluster-level evaluation:
$$\mathrm{DFS}(y)=\frac{1}{|\mathcal{A}_{y}|}\sum_{\mathcal{A}_{y_{i}}\in\mathcal{A}_{y}}\sum_{a\in\mathcal{A}_{y_{i}}}\mathbb{I}_{[a\text{ is supported by }C_{i}^{*}]}.$$

To evaluate the risk in the generator's response, Chen et al. \cite{chen2024controlling} introduced the divided cases of the generated answer, answerable(A) and unanswerble(U), along with the different prediction process in the RAG system, keep(K) and discard(D).
Four risk-aware evaluation metrics from various aspects are defined as: 

1) $Risk$ that measures the proprotion of risky casess among the kept samples:
$$Risk=\frac{|\mathrm{UK}|}{|\mathrm{AK}|+|\mathrm{UK}|}$$
2) $Carefulness$ indicates the percentage of incorrect and discarded samples that are equivalent to recall for the unanswerable samples:
$$Carefulness=\frac{|\mathrm{UD}|}{|\mathrm{UK}|+|\mathrm{UD}|}$$
3) $Alignment$ refers to the proportion of samples in which the system's judgment align with the assigned labels:
$$Alignment=\frac{|\mathrm{AK}|+|\mathrm{UD}|}{|\mathrm{AK}|+|\mathrm{AD}|+|\mathrm{UK}|+|\mathrm{UD}|}$$
4) $Coverage$ quantifies the proportion of samples retained: 
$$Coverage=\frac{|\mathrm{AK}|+|\mathrm{UK}|}{|\mathrm{AK}|+|\mathrm{AD}|+|\mathrm{UK}|+|\mathrm{UD}|}$$

\subsubsection{LLM Representation based Methods}

The representation-based methods, conversely, captures valuable metrics by modeling vector representation in the intermediate or final layers of the LLM. 
These methods can mitigate overreliance on surface lexical patterns, but they may lose interpretability since the final numeric similarity does not necessarily clarify which factual detail is correct or not.

Certain methods are inspired by the conventional metrics, demonstrated as expansions of existing metrics on the LLM. 
For instance, GPTScore \cite{fu2024gptscore} is a GPT based LLM-scoring method inspired by BertScore, which has been widely used as a convincing metric.
ARES \cite{ARES} combined a classifier with LLM embeddings to check whether a generative answer is semantically aligned with ground-truth evidence. 
RAGAS \cite{RAGAS} uses a cosine similarity approach on LLM-generated embeddings to gauge answer relevance. 

Moreover, numerous researchers have developed novel representation based metrics, which serve not only to evaluate the components but also to guide the further enhancement. 

Zhao et al. \cite{zhao2023thrust} introduced a novel metric, \textit{Thrust}, which assesses the LLM's knowledgeability by leveraging the representation distribution of the instances produced by the LLM.
A hypothesis was proposed that if an LLM has acquired adequate knowledge pertaining to a task, it should effectively cluster samples related to that task through its hidden states. 
The \textit{Thrust} metric was defined as:
$$s_{\mathrm{thrust}}(q)=\left\|\frac{1}{N\cdot K}\sum_{l=1}^N\sum_{k=1}^K\frac{|\mathcal{C}_{kl}|}{\|d_{kl}(q)\|^2}\cdot\frac{d_{kl}(q)}{\|d_{kl}(q)\|}\right\|,$$
where $N$ is the number of classes for the specific task, $K$ is the number of clusters per class, $|\mathcal{C}_{kl}|$ denotes the cardinality of the set.
$d_{kl}(q)$ is a vector pointing from the representacion of the query to the centroid.

Zhu et al. \cite{zhu2024information} introduced the information bottleneck theory into  retrieval component to messure the relevance of the recalled document and candidate document. Moreover, a new information bottleneck-based loss function was derived and used to train a better noise filter for the retriever. 
Given the sample $\{q, x, y\}$ from the dataset and the noise filter $p(\tilde{x}|x,q)$ (need tuning), the information bottleneck in the RAG task is derived and formulated as: 
$$\mathrm{IB}(\tilde{x})=P_{\mathrm{LLM}}(x|[q,\tilde{x},y])-\alpha P_{\mathrm{LLM}}(y|[q,\tilde{x}]),$$
where $[\cdot]$ means the concatenation operation. $P_{\mathrm{LLM}}$ means the final output probability of the LLM.

Li et al. \cite{li2024role} proposed a new metric $GECE$ based on METEOR for assessing the extent of the long-tailness of the generated text in RAG:
$$GECE=\frac{|\text{METEOR}(pred,ref)-\frac{1}{n}\sum_{i=1}^nP_{LLM}(t_i)|}{\alpha\cdot[E(\bigtriangledown_{ins})\cdot\bigtriangledown_{ins}]},$$
where $\alpha$ is the average word frequency, $\bigtriangledown_{ins}$ and $E(\bigtriangledown_{ins})$ are the gradient w.r.t. the current instance and the mean gradient of the total dataset, separately. 
A long-tail instance usually has a smaller $\alpha$ and $\bigtriangledown_{ins}$, obtaining a larger $GECE$, which implies larger degree of long-tailness.

To assess the extent to which external knowledge is utilized in the RAG response, Sun et al. \cite{sun2024redeep} proposed External Context Score $\mathcal{E}$, which is defined on the response level as:
$$\mathcal{E}_{\mathbf{r}}^{l,h}=\frac{1}{|\mathbf{r}|}\sum_{t\in\mathbf{r}}\mathcal{E}_{t}^{l,h}=\frac{1}{|\mathbf{r}|}\sum_{t\in\mathbf{r}}\frac{\boldsymbol{e}\cdot \boldsymbol{x}_t^L}
{\|\boldsymbol{e}\|\|\boldsymbol{x}_t^L\|},$$ 
where $|\mathbf{r}|$ means the length of the response $\mathbf{r}$, $\boldsymbol{x}_t^L$ is the $t$-th token's vector logit of the last layer $L$. 
$\boldsymbol{e}$ is a pooled vector of the most relevant vectors of $\boldsymbol{x}_t^L$ according to the attention weights in the middle layer:
$$\boldsymbol{e}=\frac{1}{|\mathcal{I}_t^{l,h}|}\sum_{j\in\mathcal{I}_t^{l,h}}\boldsymbol{x}_j^L,$$
where $\mathcal{I}_t^{l,h}$ means the attended times where the token has larger than top-k\% attention scores with $\boldsymbol{x}_t^L$ in the $l$-th layer.

Noted that some of these LLM based evaluation metrics represent research specializations. 
While they may not be directly targeted towards an actual RAG system, their presentation is an integral part of advancing researches in the field of RAG, indicating significant contributions as well.

\section{External Evaluation}
\label{external evaluation}

We have dissected the components of RAG and provided a comprehensive account of its internal evaluation. 
This section shifts our focus to \textit{the external utility that RAG, as a complete system}, encounters.
We summarize the external utility in two areas: safety and efficiency, the evaluation of whom are introduced below.

\subsection{Safety Evaluation}

Safety pertains to the RAG system's capacity to ensure the generation of stable and harmless content within a dynamic, even noisy or hazardous environment. 
As RAG systems continue widespread deployment, safety concerns have intensified beyond those of standalone LLMs. 
The incorporation of external knowledge sources introduces unique vulnerabilities requiring specialized evaluation frameworks \cite{Zhou2024a}.

\textbf{Robustness} evaluations focus on system behavior when processing misleading information in retrieval results. The RECALL benchmark \cite{RECALL} tests discrimination between reliable and counterfactual knowledge using BLEU, ROUGE-L, and specialized metrics like Misleading Rate. Wu et al. \cite{Wu2024a} quantify susceptibility to semantically related but irrelevant information using Misrepresentation Ratio and Uncertainty Ratio. SafeRAG \cite{Liang2025a} categorizes challenges like "inter-context conflict" with specific evaluation metrics, while C-RAG \cite{Kang2024} provides theoretical guarantees on generation risks using conformal risk analysis and ROUGE-L.
Cheng et al. \cite{chengxrag} introduce two metrics to evaluate the RAG system:  
1) \textit{Resilience Rate}, aiming to emphases the system's stability and robustness, quantifies the percentage of instances where the system's responses remain accurate, both prior to and following retrieval augmentation. 
2) \textit{Boost Rate} quantifies the proportion of instances initially answered erroneously that were subsequently corrected upon the introduction of a retrieved document, evaluating the effectiveness of RAG. 

\textbf{Factuality} focuses on generating accurate information and avoiding plausible but incorrect statements (hallucinations), especially with noisy or conflicting retrieval results \cite{Self-RAG,IRCoT,RECALL}. Key metrics include \textit{Factual Accuracy}, using standard QA metrics (EM, F1, accuracy, etc.) when the context might be misleading \cite{RECALL}; the \textit{Hallucination Rate}, the frequency of generated information not supported by or contradicting retrieved documents, often measured via LLM-as-judge \cite{RGB} or human evaluation; \textit{Citation Accuracy}, assessing correct attribution to sources using \textit{Citation Precision} and \textit{Citation Recall} \cite{RGB,Zhou2024a}; and \textit{Faithfulness Metrics}, evaluating how accurately the output reflects retrieved information \cite{Self-RAG}. 

\textbf{Adversarial attacks} target specific components within the RAG pipeline. Knowledge database poisoning (PoisonedRAG \cite{PoisonedRAG}) targets the retrieval corpus by injecting malicious texts that trigger predetermined outputs when retrieved.
This attack vector is evaluated using Attack Success Rate (ASR) and retrieval-focused Precision/Recall/F1 metrics. Retrieval hijacking (HijackRAG \cite{Zhang2024b}) exploits ranking algorithms to prioritize malicious content during retrieval, with evaluation focusing on attack transferability across models. Phantom attacks \cite{Chaudhari2024} use trigger-activated documents evaluated through Retrieval Failure Rate (Ret-FR), while jamming attacks \cite{Shafran2024} insert `blocker' documents that force response refusal, assessed through oracle-based metrics.

\textbf{Privacy} assess information exposure risks from retrieval databases or user queries \cite{Zeng2024}. 
Evaluation often involves simulated attacks \cite{TrojanRAG,Phantom}. 
Key metrics about privacy include the \textit{Extraction Success Rate}, the frequency or success rate of attacks extracting specific private information (e.g., names, PII) from the knowledge base, often measured by the count of successfully extracted items \cite{Zeng2024}; the \textit{PII Leakage Rate}, the amount or percentage of Personally Identifiable Information inadvertently revealed in generated outputs, typically found via pattern matching or inspection \cite{RedTeam}; and the \textit{Membership Inference Attack Success}, which measures an attacker's ability to determine if a specific data record was in the RAG system's knowledge base.

\textbf{Fairness} examines if the RAG system exhibits or amplifies biases from retrieved documents or training, leading to inequitable outputs \cite{FairRAG}. \textit{Bias Metrics} are used to analyze the outputs for disparities, which are quantitative measures of performance disparities (e.g., error rates, sentiment scores) across demographic groups \cite{FairRAG}. \textit{Stereotype Detection} measures the frequency or severity of harmful stereotypes in generated text, assessed via lists or human evaluation. \textit{Counterfactual Fairness} checks if outputs change inappropriately when sensitive attributes in queries or context are altered.

\textbf{Transparency / Accountability} assesses the understandability and traceability of the RAG system's reasoning process, enabling verification of sources and justification \cite{MetaRAG,RAG-Ex}. Metrics are often qualitative or user-focused, such as \textit{Explanation Quality}, based on human ratings of the clarity, completeness, and usefulness of explanations or provenance information \cite{RAG-Ex}; \textit{Traceability}, the ease of linking the final output back to specific source documents or passages; and \textit{Citation Accuracy} (precision/recall) \cite{Zhou2024a}.

\text{Comprehensive safety benchmarks} standardize evaluation across multiple dimensions. 
SafeRAG \cite{Liang2025a} classifies attack tasks into four categories with tailored datasets. VERA framework \cite{Ding2024} uses bootstrap sampling for confidence bounds on safety metrics, while DeepTeam's red teaming approach \cite{RedTeam} identifies vulnerabilities through systematic testing.
In addition, current research indicates defense mechanisms remain insufficient against sophisticated attacks \cite{Chaudhari2024,Zhang2024b,PoisonedRAG}. 
Evaluations reveal significant vulnerabilities in current RAG systems \cite{Chaudhari2024,Zhang2024b}, underscoring the need for robust benchmarks and metrics addressing the unique safety challenges arising from the retrieval-generation interplay.
Further efforts are required to evaluate the safety of RAG. 

\subsection{Efficiency Evaluation}
Efficiency is another crucial aspect of RAG's utility, directly linked to the real-world significance of a system's popularity, cost, and effectiveness.

\textbf{Latency} evaluation typically focuses on two critical metrics. Time to first token (TTFT) \cite{TTFT} measures the time taken by the system to produce its initial output token after receiving a query, which is particularly crucial for user experience as it directly impacts perceived responsiveness. This metric is especially important in interactive applications where immediate feedback maintains user engagement.
Additionally, complete response time (total latency) measures the duration from query submission to the generation of the entire response. This encompasses retrieval time, processing time, and generation time for all tokens. 
Hofstatte et al. \cite{EfficientRAG} proposed Single Query Latency that refers to the complete end-to-end time taken to process a single query, including both complete retrieval and generation phases.

\begin{table*}[h!]
\centering
\caption{Overview of RAG benchmarks and their evaluation datasets. Source Domain indicates the data origin (e.g., real-time news, specialized corpora), and Special Points highlight unique or novel features (like domain-specific tasks, dynamic changes, or false-premise data).}
\vspace{0.15cm}
\label{tab:dataset}
\renewcommand{\arraystretch}{1.15}
\resizebox{1.\textwidth}{!}{%
\begin{tabular}{llccc}
\toprule
\textbf{Benchmark} & \textbf{Time} & \textbf{Dataset Name(s)} & \textbf{Source Domain} & \textbf{Special Points} \\
\midrule
\rowcolor{gray!20}
RAGAS \cite{RAGAS}& 2023.09& WikiEval& Post-2022 Wikipedia& Manually labeled for faithfulness \\
FreshLLMs \cite{FreshLLMs}& 2023.11& FRESHQA& Real-time news/web queries& Dynamic QA with false-premise detection \\
\rowcolor{gray!20}
RECALL \cite{RECALL}& 2023.11& EventKG, UJ& Multilingual KGs, sci. terms& Edited/counterfactual context tests \\
ARES \cite{ARES}& 2023.11& \begin{tabular}[c]{@{}l@{}}NQ \cite{NQ}, HotpotQA \cite{HotpotQA}, FEVER \cite{FEVER},\\ WoW \cite{WoW}, MultiRC \cite{MultiRC}, ReCoRD \cite{ReCoRD}\end{tabular}& KILT and SuperGLUE corpora& Re-uses classic QA sets, multi-domain \\
\rowcolor{gray!20}
RGB \cite{RGB}& 2023.12& Custom corpus& Latest news articles& Emphasizes info integration, noise rejections \\
MultiHop-RAG \cite{MultiHop-RAG}& 2024.01& Generated corpus& Daily news segments via mediastack& Multi-hop cross-document queries \\
\rowcolor{gray!20}
CRUD-RAG \cite{CRUD-RAG}& 2024.02& \begin{tabular}[c]{@{}l@{}}Generated corpus, UHGEval\end{tabular}& Chinese news, domain texts& Create/Read/Update/Delete tasks \\
MedRAG \cite{MedRAGBench}& 2024.02& MIRAGE& Medical QA corpora& Healthcare domain knowledge \\
\rowcolor{gray!20}
FeB4RAG \cite{FeB4RAG}& 2024.02& FeB4RAG, BEIR \cite{BEIR}& Federated search tasks& Multi-domain, multi-engine retrieval \\
RAGBench \cite{RAGBench}& 2024.06& \begin{tabular}[c]{@{}l@{}}PubMedQA, CovidQA, HotpotQA,\\ MS~Marco, CUAD, DelucionQA,\\ EManual, TechQA, FinQA, TAT-QA\end{tabular}& Multi-domain corpora& Faithfulness with TRACe (Util, Rel, Adh, Compl) \\
\rowcolor{gray!20}
ReEval \cite{ReEval} & 2024.05 & {NQ (MRQA) + RealTimeQA} & Wikipedia, real-time QA & \begin{tabular}[c]{@{}c@{}}Adversarial test cases \\ for hallucination detection \end{tabular} \\
DomainRAG \cite{DomainRAG}& 2024.06& Generated admission QA& College docs with yearly updates& Single-/multi-doc, single-/multi-turn QA \\
\rowcolor{gray!20}
Telecom RAG Eval. \cite{TelecomRAGEval}& 2024.07& TeleQuAD& 3GPP-based domain docs& Triple-labeled QA from SMEs (telecom context) \\
LegalBench-RAG \cite{LegalBench-RAG}& 2024.08& \begin{tabular}[c]{@{}l@{}}PrivacyQA, CUAD, MAUD,\\ ContractNLI\end{tabular}& Expert-annotated legal corpora& Emphasizes strict retrieval of legal text \\
\rowcolor{gray!20}
RAGEval \cite{RAGEval}& 2024.08& DragonBall& Finance, law, medical docs& Schema-based generation, scenario-specific \\
CoURAGE \cite{CoURAGE}& 2024.09& \begin{tabular}[c]{@{}l@{}}RealTimeQA \cite{RealTime}, NQ \cite{NQ}\end{tabular}& Online QA + KILT tasks& Hallucination resilience, dynamic updates \\
\rowcolor{gray!20}
RAG Unfairness \cite{FairnessRAG}& 2024.09& TREC22~FairRank, BBQ& Wikipedia-based track + socioecon. QA& Fairness metrics, group disparity \\
CoFE-RAG \cite{CoFE-RAG}& 2024.10& CoFE data& PDF, DOC, multi-lingual docs& Fine-grained chunking, multi-keyword approach \\
\rowcolor{gray!20}
OCR Hinders RAG \cite{OCRRAG}& 2024.12& \begin{tabular}[c]{@{}l@{}}1,261 PDFs + 8,561 images\end{tabular}& OCR text from scanned docs& Evaluates noise from OCR errors \\

OmniEval \cite{OmniEval}& 2024.12& Finance domain set& Financial docs, numeric tasks& Emphasizes numeric correctness/factual QA \\ 
\rowcolor{gray!20}
CRAG \cite{CRAG}& 2024.12& KG + web corpus& Knowledge graphs + web pages& Multi-entity queries, curated dynamic facts \\

RAG Playground \cite{RAG-Playground}& 2024.12& 319 QA pairs& Curated multi-domain tasks& Prompt engineering / user flows \\
\rowcolor{gray!20}
MTRAG \cite{MTRAG}& 2025.01& \begin{tabular}[c]{@{}l@{}}CLAPNQ, FiQA, Govt, Cloud\end{tabular}& Wikipedia, finance, gov, tech docs& Multi-turn, bridging queries \\

CDQA \cite{CDQA}& 2025.01& Chinese~Dynamic~QA& Recent Chinese news queries& Time-varying evolving answers \\
\rowcolor{gray!20}
U-NIAH \cite{U-NIAH}& 2025.03& Starlight Academy& Synthetic “needle-in-haystack” data& Evaluates extremely long contexts \\

SCARF \cite{SCARF} & 2025.04& (User-provided) & Generic multi-domain& \begin{tabular}[c]{@{}c@{}}Modular or black-box approach \\ integrates wide metrics (LLM judge)\end{tabular} \\

\bottomrule
\end{tabular}%
}
\end{table*}

\textbf{Resources and Money Cost} evaluation of RAG systems is another critical component for assessing the efficiency. 
Cost evaluation methodologies typically focus on quantifying both direct expenditures and efficiency metrics that impact overall system economics.
The total cost of RAG systems can be categorized into several key components \cite{MeidumRAGCost}:
\begin{itemize}
    \item \textit{Infrastructure Costs}: Computing local resources for embedding generation, vector database maintenance, and LLM inference for open models.
    \item \textit{Token-based Expenses}: API charges for external LLM services based on input and output token usage.
    \item \textit{Storage Costs}: Vector database hosting and maintenance expenses that scale with corpus size.
    \item \textit{Operational Overhead}: Human supervision, system maintenance, and regular updates to knowledge bases.
    \item \textit{Development Costs}: Initial implementation, integration, and customization expenses.
\end{itemize}

For more details in the token-based expenses, LLM providers such as OpenAI and Google offer token usage metrics that track input and output token consumption during evaluation processes. This approach calculates costs by multiplying token counts by their respective pricing rates \cite{Zhang2025}.
Researchers have developed metrics to evaluate the economic efficiency of RAG implementations:

\begin{itemize}
    \item \textit{Cost-Effectiveness Ratio}: Measures performance improvement per unit of cost, allowing for standardized comparison between different RAG configurations \cite{Zhang2025}.
    \item \textit{Retrieval Precision ROI}: Quantifies the economic return of improving retrieval precision by measuring the reduction in irrelevant context processing costs \cite{Zhang2025}. This metric demonstrated that optimizing retrieval can improve cost efficiency by up to around 50\% through reducing token consumption during LLM inference.
    \item \textit{User-Controllable Cost-Accuracy Tradeoffs}: Su et al. \cite{Su2025} propose evaluation methods using an interpretable control parameter ($\alpha$) that allows systematic assessment of the relationship between retrieval costs and accuracy. This approach enables evaluating RAG systems across a spectrum of cost constraints rather than at fixed operating points.
    \item \textit{Comparative Cost Analysis}: Methodologies for evaluating relative cost efficiency between different RAG implementations for specific use cases, considering both direct costs and long-term economic sustainability \cite{Sakar2025}.
\end{itemize}




\begin{table*}[t!]
\centering
\caption{RAG evaluation frameworks, highlighting principal evaluation targets and methods. Retrieval focuses mainly on \textcolor[HTML]{9A0000}{Relevance (R)}, \textcolor[HTML]{CD9934}{Correctness (C)} or  \textcolor[HTML]{9AF000}{Comprehensiveness}, whereas generation (right) focuses on \textcolor[HTML]{F8A102}{Faithfulness (F)}, \textcolor[HTML]{036400}{Correctness (C)}, or \textcolor[HTML]{6665CD}{Relevance (R)}. External evaluation targets (\emph{safety}, \emph{efficiency}) or other statements appear in italics. }
\vspace{0.125cm}
\label{table:rag-frameworks-complete}
\renewcommand{\arraystretch}{1.0}
\resizebox{0.97\textwidth}{!}{%

\begin{tabular}{llcccc}
\toprule
\textbf{Type} & \textbf{Framework} & \textbf{Time} & \textbf{Raw Targets} & \textbf{Retrieval Metrics} & \textbf{Generation Metrics} \\
\midrule
\rowcolor{gray!20}
Research& FiD-Light \cite{EfficientRAG}& 2023.07& \textit{Latency}& --& -- \\
Research& Diversity Reranker \cite{HaystackDiversity}& 2023.08& \textcolor[HTML]{9AF000}{Diversity}& \textcolor[HTML]{9AF000}{Cosine Distances}& -- \\
\rowcolor{gray!20}
Benchmark& RAGAS \cite{RAGAS}& 2023.09& \begin{tabular}[c]{@{}c@{}}\textcolor[HTML]{9A0000}{Context R}, \textcolor[HTML]{6665CD}{Answer R}, \textcolor[HTML]{F8A102}{F}\end{tabular}& \textcolor[HTML]{9A0000}{LLM as Judge}& \begin{tabular}[c]{@{}c@{}}\textcolor[HTML]{6665CD}{LLM CosSim},\\ \textcolor[HTML]{F8A102}{LLM as Judge}\end{tabular} \\
Tool& TruEra RAG Triad \cite{TruEra}& 2023.10& \begin{tabular}[c]{@{}c@{}}\textcolor[HTML]{9A0000}{Context R}, \textcolor[HTML]{6665CD}{Answer R}, \textit{Groundedness}\end{tabular}& \textcolor[HTML]{9A0000}{LLM as Judge}& \textcolor[HTML]{6665CD}{LLM as Judge} \\
\rowcolor{gray!20}
Tool& LangChain Bench.\ \cite{LangChain2023}& 2023.11& \begin{tabular}[c]{@{}c@{}}\textcolor[HTML]{CD9934}{C}, \textcolor[HTML]{F8A102}{F}, \textit{ExecutionTime}, \textit{EmbCosDist}\end{tabular}& \textcolor[HTML]{CD9934}{Exact-match}& \textcolor[HTML]{F8A102}{LLM as Judge} \\
Benchmark& FreshLLMs \cite{FreshLLMs}& 2023.11& \begin{tabular}[c]{@{}c@{}}\textcolor[HTML]{036400}{Response C}, \textit{Fast-changing}, \textit{False premise}\end{tabular}& \textcolor[HTML]{CD9934}{(retrieval logs)}& \begin{tabular}[c]{@{}c@{}}\textcolor[HTML]{036400}{STRICT / RELAXED},\\ \textit{FRESHEVAL} (LLM-based)\end{tabular} \\
\rowcolor{gray!20}
Tool& RECALL \cite{RECALL}& 2023.11& \begin{tabular}[c]{@{}c@{}}\textcolor[HTML]{036400}{Response Quality}, \textit{Robustness}\end{tabular}& --& \textcolor[HTML]{036400}{BLEU, ROUGE-L} \\
Benchmark& ARES \cite{ARES}& 2023.11& \begin{tabular}[c]{@{}c@{}}\textcolor[HTML]{9A0000}{Context R}, \textcolor[HTML]{F8A102}{Answer F}, \textcolor[HTML]{6665CD}{Answer R}\end{tabular}& \textcolor[HTML]{9A0000}{LLM + Classifier}& \begin{tabular}[c]{@{}c@{}}\textcolor[HTML]{F8A102}{LLM + Classifier},\\ \textcolor[HTML]{6665CD}{LLM + Classifier}\end{tabular} \\
\rowcolor{gray!20}
Benchmark& RGB \cite{RGB}& 2023.12& \begin{tabular}[c]{@{}c@{}}\textcolor[HTML]{F8A102}{Info Integration}, \textit{NoiseRobust}, \\ \textit{NegRejection}, \textit{Counterfact}\end{tabular}& --& \textcolor[HTML]{F8A102}{Accuracy} \\
Tool& Databricks Eval \cite{DatabricksRAGEval}& 2023.12& \begin{tabular}[c]{@{}c@{}}\textcolor[HTML]{036400}{C}, \textit{Readability}, \textcolor[HTML]{9AF000}{Comprehensiveness}\end{tabular}& --& {LLM as Judge} \\
\rowcolor{gray!20}
Benchmark& MultiHop-RAG \cite{MultiHop-RAG}& 2024.01& \begin{tabular}[c]{@{}c@{}}\textcolor[HTML]{CD9934}{Retrieval C}, \textcolor[HTML]{036400}{Response C}\end{tabular}& \textcolor[HTML]{CD9934}{MAP, MRR, Hit@K}& \textcolor[HTML]{036400}{LLM as Judge} \\
Benchmark& CRUD-RAG \cite{CRUD-RAG}& 2024.02& \textit{Create, Read, Update, Delete}& --& \begin{tabular}[c]{@{}c@{}}\textcolor[HTML]{036400}{ROUGE, BLEU},\\ \textcolor[HTML]{F8A102}{RAGQuestEval}\end{tabular} \\
\rowcolor{gray!20}
Benchmark& MedRAG \cite{MedRAGBench}& 2024.02& \textcolor[HTML]{036400}{Accuracy (medical)}& --& \textcolor[HTML]{036400}{Exact-match, Acc.} \\
Benchmark& FeB4RAG \cite{FeB4RAG}& 2024.02& \begin{tabular}[c]{@{}c@{}}\textcolor[HTML]{F8A102}{Consistency}, \textcolor[HTML]{036400}{C}, \textit{Clarity}, \textit{Coverage}\end{tabular}& --& \begin{tabular}[c]{@{}c@{}}\textcolor[HTML]{F8A102}{Human Eval},\\ \textcolor[HTML]{036400}{Human Eval}\end{tabular} \\
\rowcolor{gray!20}
Benchmark& Arabic RAG Eval. \cite{ArabicRAGEval}& 2024.05& \begin{tabular}[c]{@{}c@{}}\textcolor[HTML]{9A0000}{Doc R}, \textcolor[HTML]{6665CD}{Answer R}\end{tabular}& \textcolor[HTML]{9A0000}{nDCG, MRR, mAP}& \textcolor[HTML]{6665CD}{Possibly CosSim to QA} \\
Benchmark& RAGBench \cite{RAGBench}& 2024.06& \begin{tabular}[c]{@{}c@{}}\textcolor[HTML]{9A0000}{Context R}, \textcolor[HTML]{6665CD}{Answer R}, \textcolor[HTML]{F8A102}{Explainability}, \\ \textit{TRACe = Util, Rel, Adh, Compl.}\end{tabular}& \textcolor[HTML]{9A0000}{LLM-based Eval}& \begin{tabular}[c]{@{}c@{}}\textcolor[HTML]{6665CD}{LLM-based Eval}, \textcolor[HTML]{F8A102}{TRACe Metrics}\end{tabular} \\
\rowcolor{gray!20}
Benchmark & ReEval \cite{ReEval} & 2024.05 &\begin{tabular}[c]{@{}c@{}}\textcolor[HTML]{F8A102}{Hallucination}\\ \textit{Adversarial Attack}\end{tabular} &-- &\begin{tabular}[c]{@{}c@{}}\textcolor[HTML]{F8A102}{F1, EM, Entailment}\\ \textcolor[HTML]{F8A102}{LLM or Human Eval}\end{tabular} \\
Benchmark& DomainRAG \cite{DomainRAG}& 2024.06& \begin{tabular}[c]{@{}c@{}}\textcolor[HTML]{036400}{C}, \textcolor[HTML]{F8A102}{F}, \textit{NoiseRobust}, \textit{StructOutput}\end{tabular}& --& \begin{tabular}[c]{@{}c@{}}\textcolor[HTML]{036400}{F1, EM}, \\\textcolor[HTML]{F8A102}{ROUGE-L}, LLM\end{tabular} \\
\rowcolor{gray!20}
Benchmark& CoURAGE \cite{CoURAGE}& 2024.06& \textcolor[HTML]{F8A102}{Hallucination}& --& \begin{tabular}[c]{@{}c@{}}\textcolor[HTML]{F8A102}{F1, EM, LLM as Judge}, \\ \textcolor[HTML]{F8A102}{Human Eval}\end{tabular} \\
Tool& Telecom RAG Eval. \cite{TelecomRAGEval}& 2024.07& \begin{tabular}[c]{@{}c@{}}\textcolor[HTML]{9A0000}{Context R}, \textcolor[HTML]{F8A102}{Faithfulness}, \\ \textcolor[HTML]{036400}{Correctness}\end{tabular}& \textcolor[HTML]{9A0000}{LLM-based Metrics}& \begin{tabular}[c]{@{}c@{}}\textcolor[HTML]{F8A102}{RAGAS-based}, \textcolor[HTML]{036400}{LLM Eval}\end{tabular} \\
\rowcolor{gray!20}
Benchmark& LegalBench-RAG \cite{LegalBench-RAG}& 2024.08& \begin{tabular}[c]{@{}c@{}}\textcolor[HTML]{9A0000}{Doc-level Precision}, \textcolor[HTML]{6665CD}{Citation Rel.}\end{tabular}& \begin{tabular}[c]{@{}c@{}}\textcolor[HTML]{9A0000}{Precision}, \textcolor[HTML]{6665CD}{Recall}\end{tabular}& -- \\
Benchmark& RAGEval \cite{RAGEval}& 2024.08& \begin{tabular}[c]{@{}c@{}}\textcolor[HTML]{F8A102}{Completeness}, \textcolor[HTML]{036400}{Hallucination}, \\ \textit{Irrelevance}\end{tabular}& LLM-based Scoring& \begin{tabular}[c]{@{}c@{}}\textcolor[HTML]{036400}{LLM-based}, \textcolor[HTML]{F8A102}{Human Alignment}\end{tabular} \\
\rowcolor{gray!20}
Benchmark& RAG Unfairness \cite{FairnessRAG}& 2024.09& \begin{tabular}[c]{@{}c@{}}\textit{Fairness}, \textcolor[HTML]{CD9934}{C}, \textcolor[HTML]{036400}{C}\end{tabular}& \textcolor[HTML]{CD9934}{MRR@K}& \begin{tabular}[c]{@{}c@{}}\textcolor[HTML]{036400}{EM, ROUGE}\end{tabular} \\
Benchmark& CoFE-RAG \cite{CoFE-RAG}& 2024.10& \begin{tabular}[c]{@{}c@{}}\textcolor[HTML]{CD9934}{Fine-grained Retrieval}, \textcolor[HTML]{F8A102}{Resp Quality}, \\ \textcolor[HTML]{9A0000}{Diversity}\end{tabular}& \begin{tabular}[c]{@{}c@{}}\textcolor[HTML]{CD9934}{Recall, Correctness}, \\\textcolor[HTML]{9A0000}{Multi-keyword}\end{tabular}& \begin{tabular}[c]{@{}c@{}}\textcolor[HTML]{F8A102}{BLEU, ROUGE-L}, \textcolor[HTML]{F8A102}{LLM as Judge}\end{tabular} \\
\rowcolor{gray!20}
Benchmark& Toward Instr.-Following \cite{InstructRAG}& 2024.10& \begin{tabular}[c]{@{}c@{}}\textcolor[HTML]{6665CD}{Instr.\ Relevance}, \textcolor[HTML]{036400}{Constraint}\end{tabular}& --& \begin{tabular}[c]{@{}c@{}}\textcolor[HTML]{6665CD}{LLM as Judge}, \\\textcolor[HTML]{036400}{Atomic Pass Rate}\end{tabular} \\
Benchmark& OmniEval \cite{OmniEval}& 2024.12& \begin{tabular}[c]{@{}c@{}}\textcolor[HTML]{F8A102}{Factual Acc.}, \textit{Domain Tasks}\end{tabular}& \textit{Rule+LLM}& \begin{tabular}[c]{@{}c@{}}\textcolor[HTML]{F8A102}{Manual or LLM FT}\end{tabular} \\
\rowcolor{gray!20}
Benchmark& CRAG \cite{CRAG}& 2024.12& \begin{tabular}[c]{@{}c@{}}\textcolor[HTML]{F8A102}{Accuracy}, \textit{Dynamism}, \textit{Complex Facts}, \\ \textcolor[HTML]{9A0000}{R}, \textcolor[HTML]{036400}{C}\end{tabular}& Weighted scoring& \textcolor[HTML]{F8A102}{Accuracy}, \textcolor[HTML]{036400}{Truthfulness measure} \\
Benchmark& OCR Hinders RAG \cite{OCRRAG}& 2024.12& \begin{tabular}[c]{@{}c@{}}\textcolor[HTML]{036400}{Accuracy}, \textit{OCR Noise}, \\ \textit{Semantic vs.\ Format Noise}\end{tabular}& \begin{tabular}[c]{@{}c@{}}\textit{EditDist}, \textit{LCS}\end{tabular}& \textcolor[HTML]{036400}{F1-score} \\
\rowcolor{gray!20}
Benchmark& RAG Playground \cite{RAG-Playground}& 2024.12& \begin{tabular}[c]{@{}c@{}}\textcolor[HTML]{CD9934}{Retrieval Strategy}, \\ \textcolor[HTML]{F8A102}{Prompt Eng.}\end{tabular}& \textcolor[HTML]{CD9934}{Comparison-based}& \textcolor[HTML]{F8A102}{LLM-based Eval} \\
Benchmark& MTRAG \cite{MTRAG}& 2025.01& \begin{tabular}[c]{@{}c@{}}\textcolor[HTML]{CD9934}{Multi-turn Quality}, \textcolor[HTML]{036400}{Conv.\ C}\end{tabular}& \textcolor[HTML]{CD9934}{Recall, nDCG}& \textcolor[HTML]{036400}{LLM as Judge} \\
\rowcolor{gray!20}
Benchmark& CDQA \cite{CDQA}& 2025.01& \textcolor[HTML]{036400}{Accuracy}& --& \textcolor[HTML]{036400}{F1} \\
Benchmark& U-NIAH \cite{U-NIAH}& 2025.03& \begin{tabular}[c]{@{}c@{}}\textcolor[HTML]{CD9934}{Needle Detect}, \textit{LongContext}, \textcolor[HTML]{036400}{No Halluc.}\end{tabular}& \textcolor[HTML]{CD9934}{Recall}& \begin{tabular}[c]{@{}c@{}}\textcolor[HTML]{036400}{LLM Judge}, Heatmap\end{tabular} \\
\rowcolor{gray!20}
Tool& eRAG \cite{eRAG}& 2024.04& \begin{tabular}[c]{@{}c@{}}\textit{Doc-level Rel.}, \textcolor[HTML]{036400}{Downstream Quality}\end{tabular}& \textit{Doc-level LLM}& \textcolor[HTML]{036400}{Kendall's $\tau$} \\
Tool & SCARF \cite{SCARF} & 2025.04 &\begin{tabular}[c]{@{}c@{}}\textcolor[HTML]{9A0000}{Context R}, \textcolor[HTML]{6665CD}{Answer R}, \\ \textcolor[HTML]{F8A102}{Faithfulness}\end{tabular} &\textcolor[HTML]{9A0000}{  \begin{tabular}[c]{@{}c@{}}LLM-based or \\ BLEU/ROUGE \end{tabular}} &\begin{tabular}[c]{@{}c@{}}\textcolor[HTML]{6665CD}{RAGAS-like Relevance},\\ \textcolor[HTML]{F8A102}{LLM-based}\\(Black-box Integration)\end{tabular} \\
\bottomrule
\end{tabular}
}
\end{table*}

\section{Resources}
\label{sec:resources}
The evaluation methodologies previously examined are comprehensive, though not necessarily abundant. This section systematically compiles, categorizes, and presents the implemented RAG evaluation frameworks, benchmarks, analytical tools, and datasets that have emerged in the large language model era. To our knowledge, this compilation constitutes the most exhaustive collection of RAG evaluation frameworks currently documented in the literature.

\textbf{Datasets}. 
We compiled the benchmarks along with the associated datasets in recent years. 
Early works focus on static general-purpose QA datasets (e.g., NQ \cite{NQ}, HotpotQA \cite{HotpotQA}), providing well-established baselines but lack recency or domain specificity. 
Recent benchmarks counter these limitations by 1) sourcing live news or rapidly updated online documents (e.g., RGB \cite{RGB}, MultiHop-RAG \cite{MultiHop-RAG}) to test time-sensitive capabilities; 2) curating domain-specific corpora in law, healthcare, or finance (e.g.,  MedRAG \cite{MedRAGBench}, OmniEval \cite{OmniEval}, LegalBench-RAG \cite{LegalBench-RAG}); or 3) generating synthetic data or specialized QA pairs, possibly with false-premise or counterfactual elements (e.g., FreshLLMs \cite{FreshLLMs}, RAGEval \cite{RAGEval}) to assess robustness and misinformation handling.
We further provide a concise description of the original domains and characteristics according to the original resource, as shown in Table \ref{tab:dataset}.
Noted that only the datasets containing retrieved ground truth documents are included, indicating a concern for more in-depth system component evaluation. 

\textbf{Frameworks with Evaluation Methods}. 
We compiled and summarized the evaluation methods devised by existing frameworks, as illustrated in Table \ref{table:rag-frameworks-complete}. 
These efforts span from initial, point-level researches \cite{EfficientRAG,HaystackDiversity} to later, multi-component evaluation tools and benchmarks \cite{LangChain2023,ARES}, encompassing a remarkably comprehensive collection of assessment frameworks.
The evaluation methods employed are varied, encompassing both traditional \cite{RECALL,ArabicRAGEval} and LLM-based metrics \cite{CRUD-RAG,RAGBench}.
Additionally, there are frameworks that facilitate safety-focused evaluations \cite{CoURAGE,RGB}, or are tailored to specific downstream domains like document \cite{OCRRAG,U-NIAH}, telecom \cite{TelecomRAGEval}, medicine \cite{MedRAGBench}, etc. 
Referencing the component evaluation objectives outlined in section \ref{sec3.1}, we categorize and highlight the evaluation elements and specific metrics.

\section{Discussion}
\label{analysis}

\subsection{Statistics and Analysis of RAG Evaluation}

The proliferation of LLM has contributed to a significant diversification of RAG evaluation methods. 
Current researches, while demonstrating comprehensive coverage of RAG evaluation dimensions, often subjectively assert their respective utility statements.
To assess the popularity of these evaluation methods, we conducted a statistical analysis of the available methods from a survey perspective. 
This can also be viewed as a research-oriented simple meta-evaluation. 
We crawled the collection of the papers since 2022 autumn with keywords about RAG in the accepted papers of the high-level conferences about NLP \& AI, and extracted the component as well as the evalauation metrics the papers focus and utilize.
We finally amassed a total of 582 PDF manuscripts.
All the included papers have undergone rigorous peer review, demonstrating scholarly merit with complete experimental methodologies and logically structured evaluation procedures.

\begin{figure}[t]
    \centering
    \includegraphics[width=.7\linewidth]{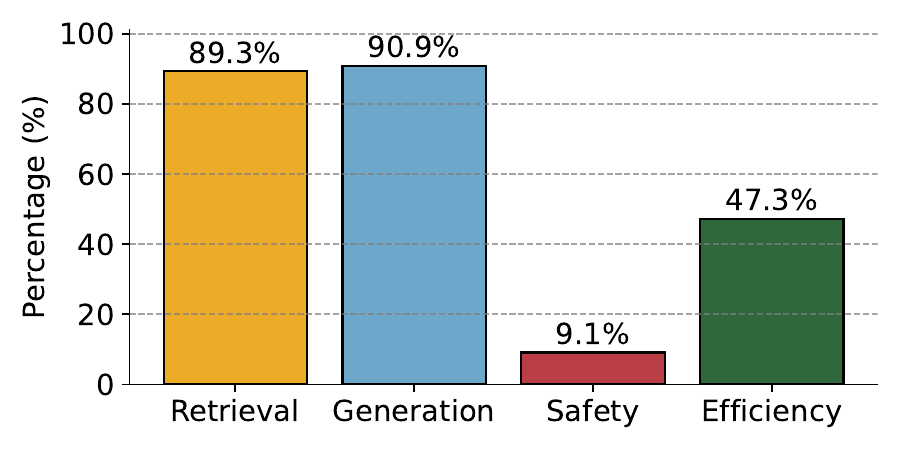}
    \caption{Statistics on the distribution of RAG studies across four key areas: retrieval, generation, safety, and efficiency. A paper may utilize evaluation methods in more than one areas. }
    \label{fig:bar}
\end{figure}

\textbf{Research Focus}. 
Figure \ref{fig:bar} illustrates the statistical distribution of evaluation methods used across the four different segments in RAG studies (Retrieval / Generation / Safety / Efficiency).
The data suggests a prevailing focus on internal research and evaluation of RAG systems, as indicated by the extensive coverage of the retrieval and generation processes.
In contrast, external evaluations, particularly those related to safety, have garnered less attention. 

\begin{figure}[t]
    \centering
    \includegraphics[width=1\linewidth]{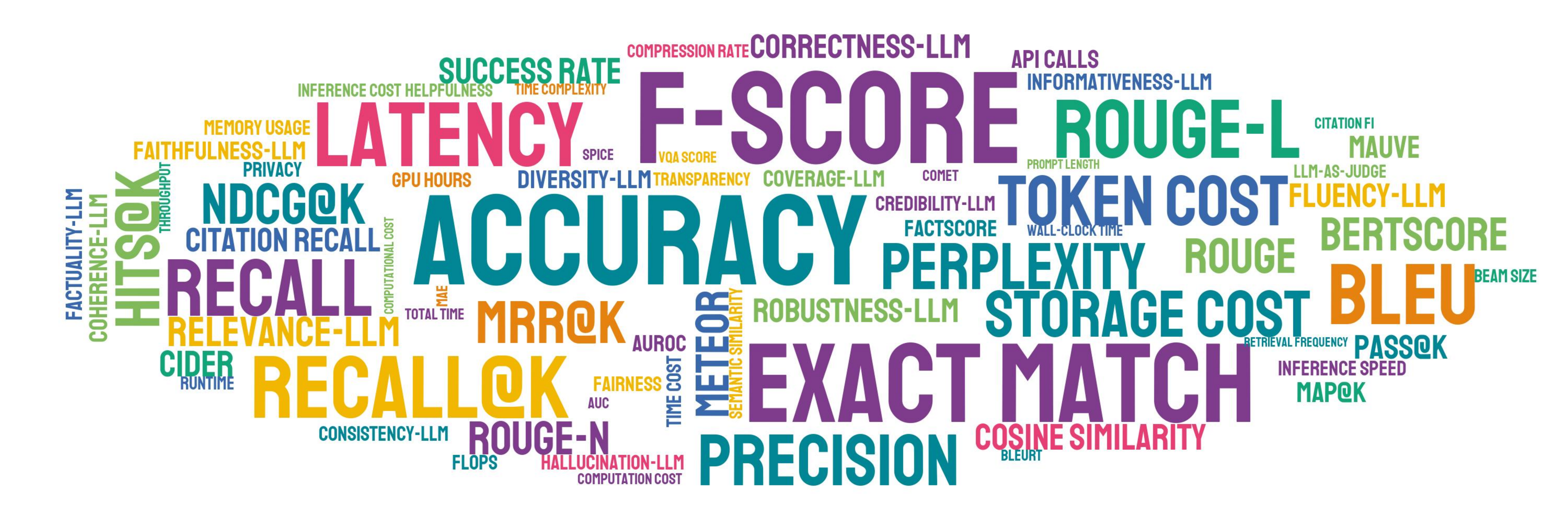}
    \caption{Frequency statistics wordcloud of evaluation metrics in RAG studies. The LLM-based methods are categorized based on the targets and presented with the suffix `-LLM'.
    F-score refers to the expanded F1-score. }
    \label{fig:cloud}
\end{figure}
\begin{figure}[t]
    \centering
    \includegraphics[width=.8\linewidth]{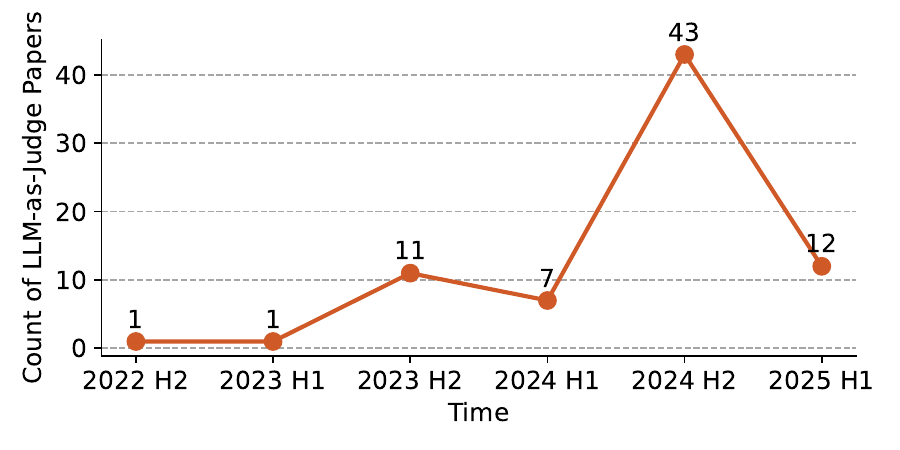}
    \caption{The number of papers explicitly mentioning LLM-based evaluation on RAG. 
    The 2025 H1 collection is up to March 31st.}
    \label{fig:llm-use}
\end{figure}

\textbf{Metric Preference}. 
Word frequency counts were conducted for the assessment metrics mentioned in the papers, with the wordcloud displayed in Figure \ref{fig:cloud}. 
Whenever a metric is formally introduced in the body of a paper or reported in the table of experimental results, its word frequency count is set +1. 
We manually merged and mapped synonymous metrics in the same session and excluded the words with global occurrences lower than twice. 
It is observed that traditional metrics predominantly dominate the evaluation usage, while LLM-based methods have not yet gained widespread acceptance among researchers. 
This phenomenon is attributed to the simplicity and reliability of the conventional metrics. 
Conversely, the LLM-based methods often require more effort and involve multiple settings that are difficult to keep the same across different researches, such as the LLM version and prompt design. 

\textbf{Trend of LLM Usage}.
Despite the potential issues with LLM-based methods, there is an observable trend of increasing application, as shown in Figure \ref{fig:llm-use}.
2024 H2 and 2025 H1 have the top two highest numbers. 
LLM judges are ultimately capable of handling more complex designs, drawing closer to real-world applications.
LLM itself, additionally,  has continued to evolve in recent years, with the performance progressively improving, and the supported functions expanding.

\subsection{Challenges and Future Directions}

This section addresses several challenges inherent in contemporary RAG evaluation. 

\textbf{Limitations of LLM-based Methods}.
The current evaluation design does not sufficiently address the timeliness and the black-box nature inherent in the LLM. 
The method of employing LLMs for assessments, particularly through direct prompts, raises latent risk about stability and security. 
Future research should focus on enhancing the robustness of the evaluation process itself and minimizing the likelihood of LLM errors in the RAG system.

\textbf{Cost of Evaluation}. 
The cost associated with the RAG system has garnered attention. 
Nevertheless, a thorough evaluation remains expensive due to the vast scale of the tools and datasets involved. 
Determining an efficient method for system evaluation, or striking a balance between cost and effectiveness, is one of the directions for future research.

\textbf{Advanced Evaluation Methods}.
As LLMs continue to evolve, the components of RAG systems are becoming more diverse. 
Currently, many of these components are evaluated using end-to-end RAG ontology metrics, with a lack of comprehensive functional decomposition evaluation or theoretical analysis. 
Concurrently, there remains untapped potential in the functionalities of LLMs themselves.
For instance, the evaluation about deep thinking models (e.g. openai-o1 \cite{jaech2024openai}) along with the thinking process of LLMs in conjunction with RAG's retrieval and generation process, is still inadequate. 
These in-depth evaluation strategies require further research and development in the future.

\textbf{Comprehensiveness of the Evaluation Framework}.
Despite the abundant evaluation frameworks at present, individual ones are somewhat limited in their metrics and methods of evaluation. 
Moreover, most contemporary frameworks concentrate on widely used languages such as English and Chinese. 
There is an urgent need for frameworks that are not only methodologically but also linguistically diverse.

\section{Conclusion}
In this paper, we have presented the first comprehensive survey of RAG evaluation methodologies in the LLM era. Our systematic analysis reveals several important insights for researchers and practitioners working with these increasingly prevalent systems.
For the evaluation of internal RAG performance, we dissect the internal components of RAG systems, define the assessment objectives, and gather a range of methods and metrics from traditional to innovative.
Moreover, we investigate the external evaluation related to system integrity such as safety and efficiency, which are underexplored in RAG research according to our statistical analysis.
Additionally, we compile and categorize the current evaluation datasets and frameworks to elucidate the unique attributes and assessment focuses of the resources.
Last but not least, we analyze the implementation of existing evaluation methods and synthesize the challenges and future directions of RAG evaluation in the LLM era.

\begin{acknowledgement}

\end{acknowledgement}

\begin{competinginterest}
The authors declare that they have no competing interests or financial conflicts to disclose. 
\end{competinginterest}

\bibliographystyle{fcs}
\bibliography{custom,custom-2,peter-benchmark,peter-verified,peter-update}


\begin{biography}{photo/Aoran_Gan.jpg}
    Aoran Gan is working toward the PhD degree in the School of Artificial Intelligence and Data Science, University of Science and Technology of China. His research interests include text mining, knowledge graph and large language models.
\end{biography}

\begin{biography}{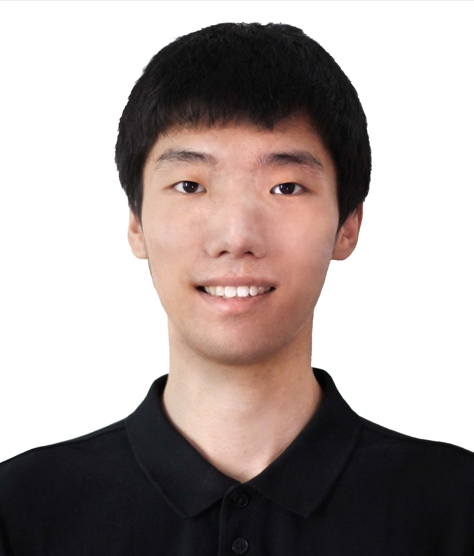}
    Hao Yu is pursuing a MS degree at McGill University and is affiliated with Quebec Artificial Intelligence Institute. His research focuses on multilingual and low-resource NLP, as well as RAG systems for misinformation detection.
\end{biography}

\begin{biography}{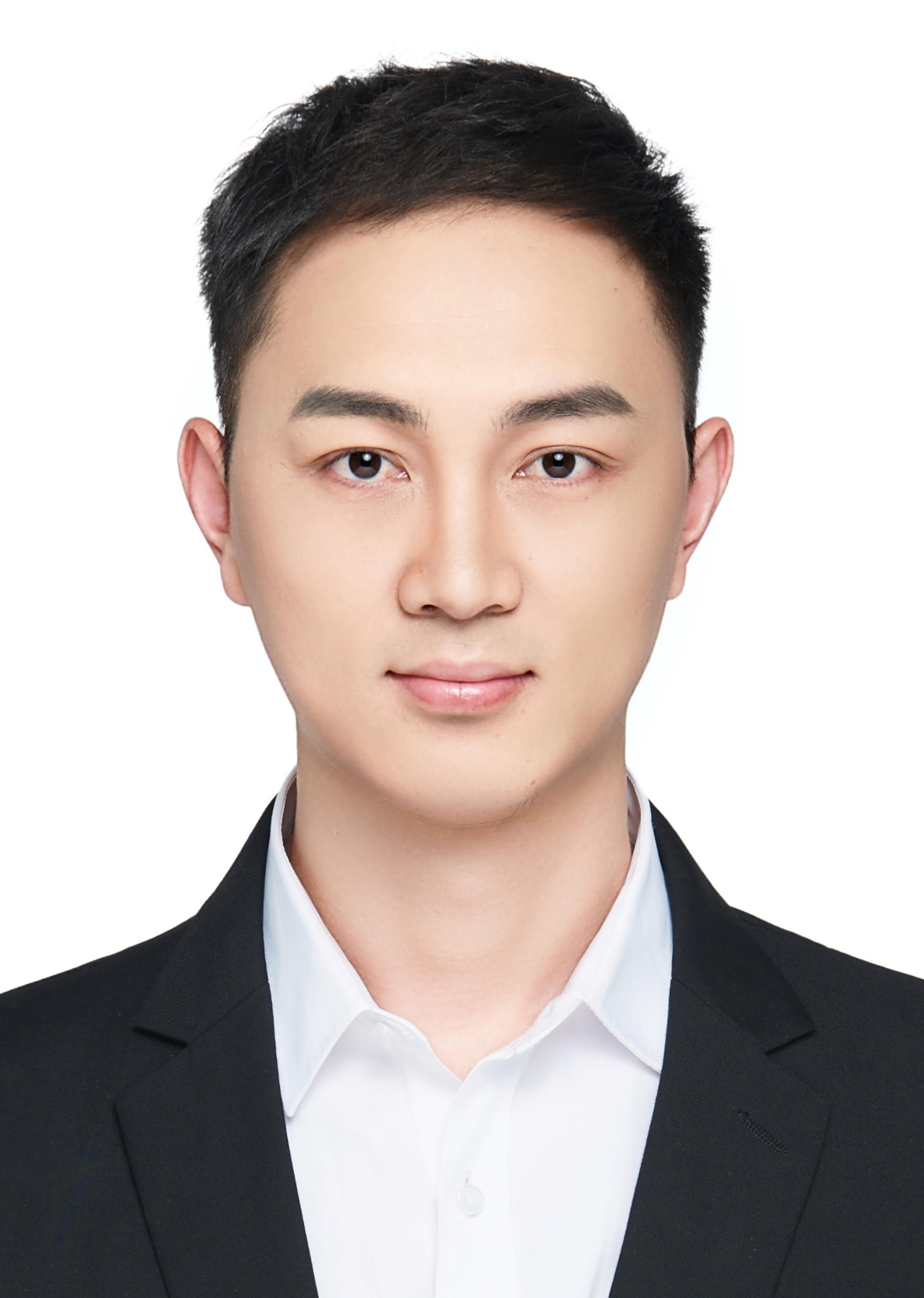}
Kai Zhang is an Associate Researcher at the University of Science and Technology of China. 
His general area of research is natural language processing and knowledge discovery. 
He is a member of ACM, SIGIR, AAAI, and CCF. 
\end{biography}

\begin{biography}{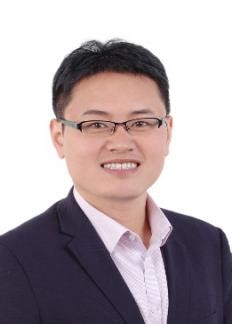}
Qi Liu is a professor in the School of Artificial Intelligence and Data Science at USTC. His area of research is data mining and knowledge discovery. He has published prolifically in refereed journals and conferences. He is an Associate Editor of IEEE TBD and Neurocomputing. 
\end{biography}

\begin{biography}{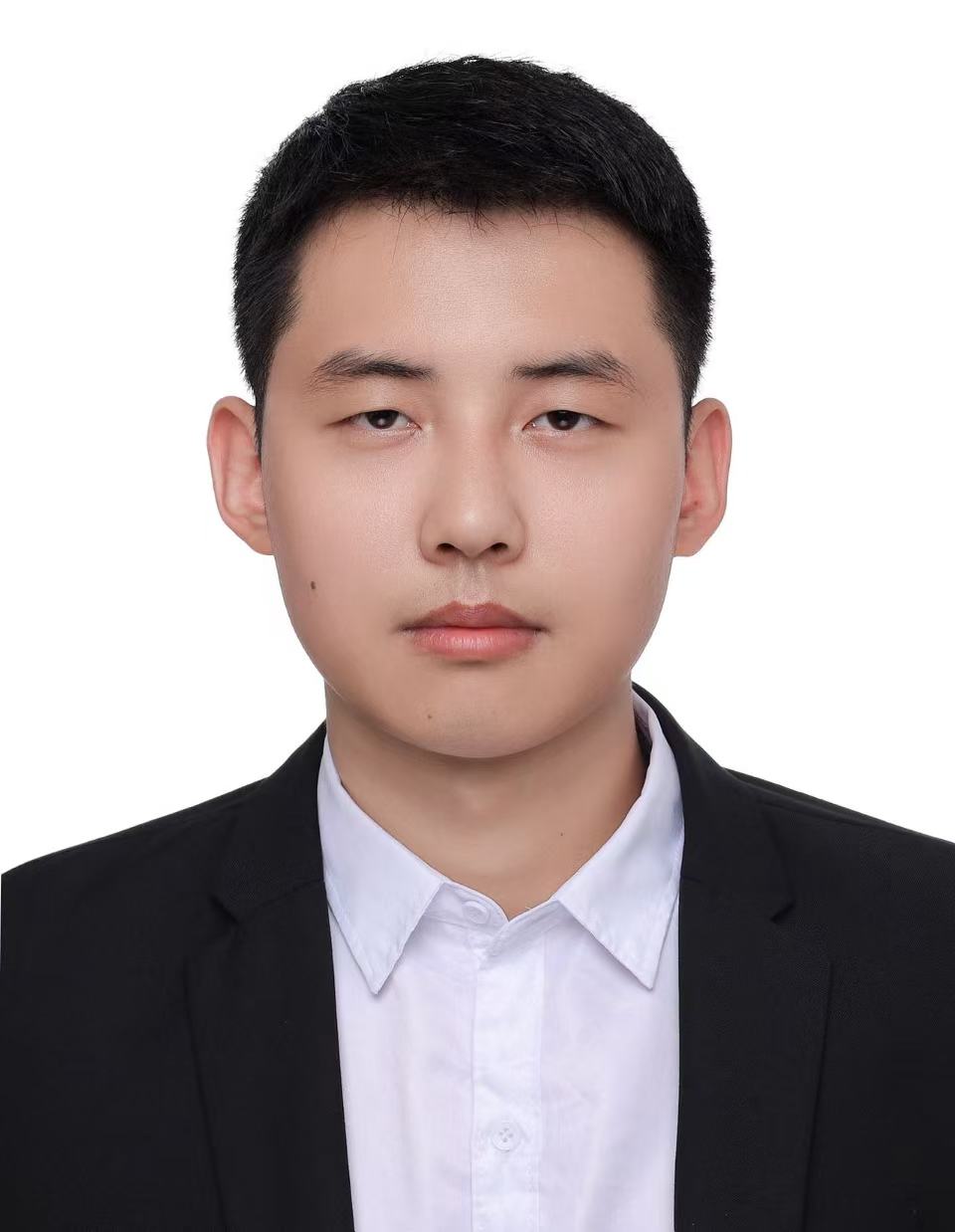}
    Wenyu Yan is currently pursuing MS degree in University of Science and Technology of China. His research interests focus on conversational search, retrieval-augmented generation, etc.
\end{biography}

\begin{biography}{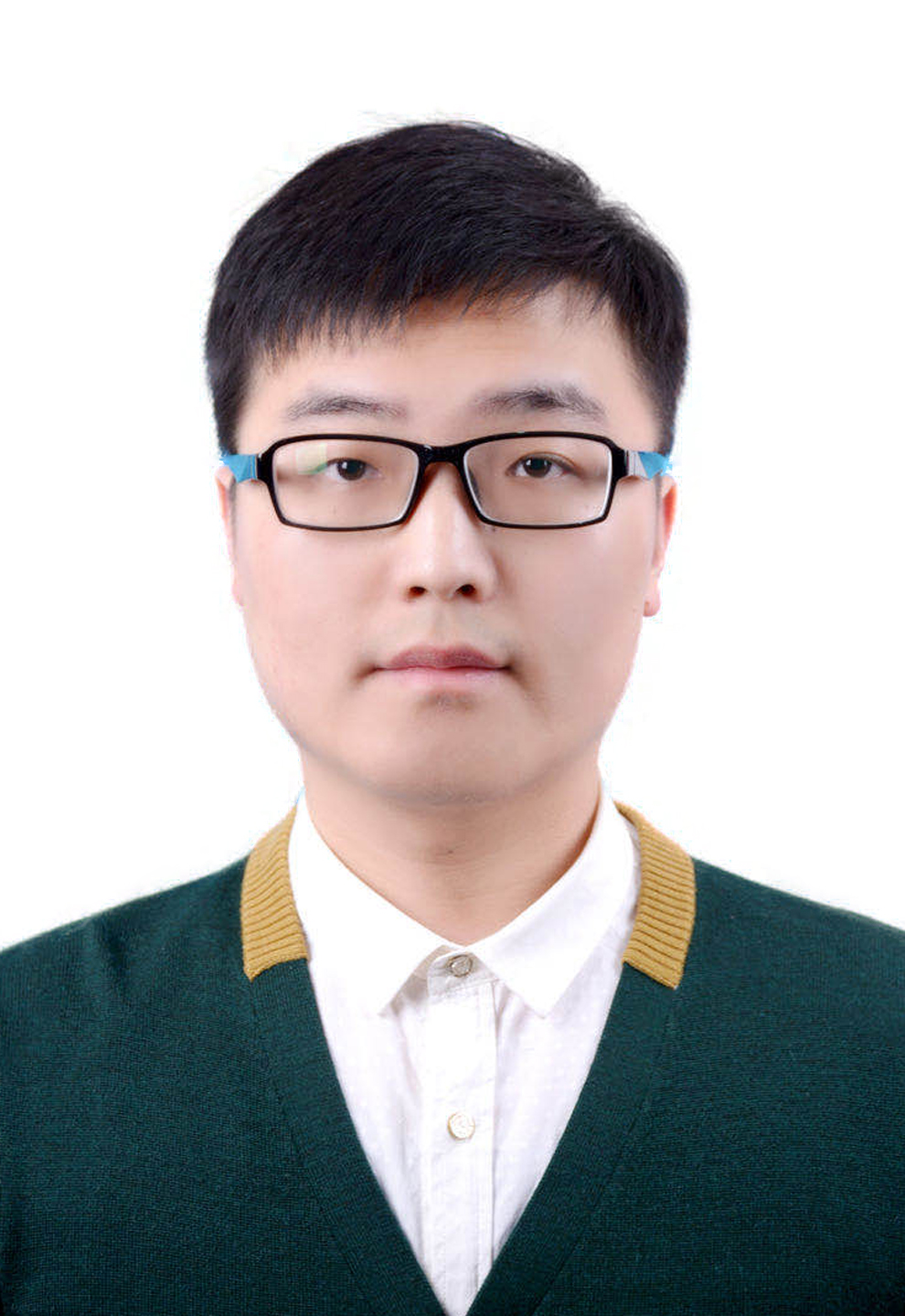}
Zhenya Huang is currently an Associate Professor with USTC. His main research interests include data mining, knowledge reasoning, natural language processing, and intelligent education. He has published more than 50 papers in refereed journals and conference proceedings.
\end{biography}

\begin{biography}{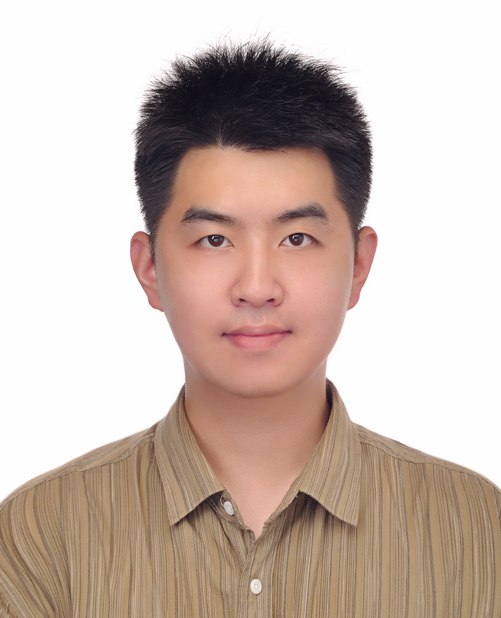}
Shiwei Tong is a senior data scientist at Tencent Company. His research focuses on Game Data Mining and Game Applications driven by Large Language Models.
 \end{biography}

\begin{biography}{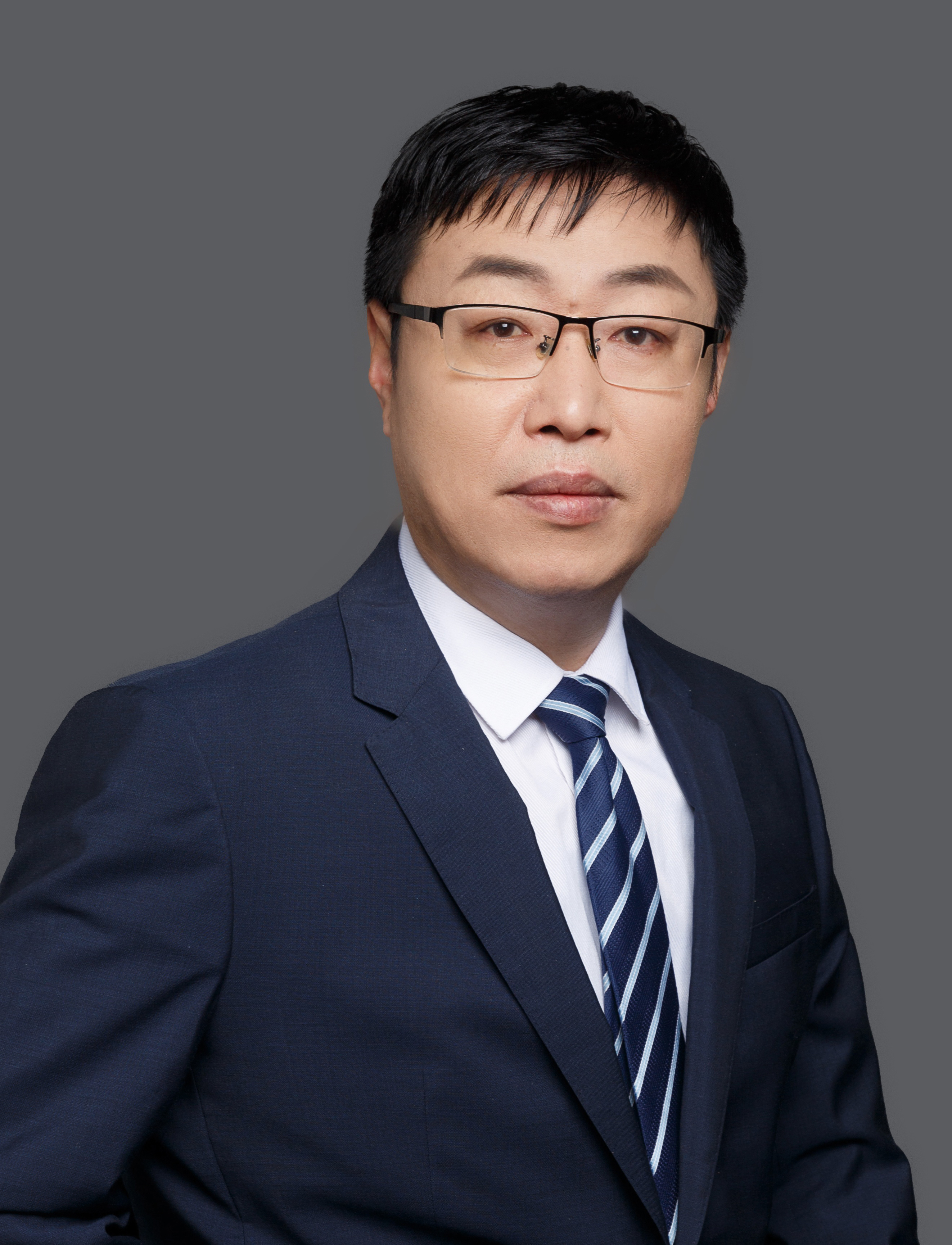}
Enhong Chen is a professor in the
School of Computer Science and Technology at
USTC. His general area of research includes data mining and machine learning, social network analysis, and recommender systems. He was on program committees of numerous conferences including SIGKDD, ICDM, and SDM.
\end{biography}

\begin{biography}{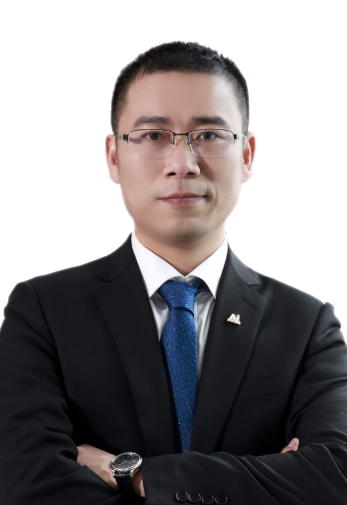}
    Guoping Hu is senior vice president of iFLYTEK, director of the National Key Laboratory of Cognitive Intelligence.
    He has been honored with the First Prize of State Science and Technology Advancement Award and garnered over 300 authorized patents. 
\end{biography}

\end{document}